\documentclass[twoside]{article}
\usepackage[accepted]{aistats2018}
\usepackage{amsmath}
\usepackage{amssymb}
\usepackage{natbib}
\usepackage{graphicx}
\usepackage{nicefrac}

\usepackage{color}
\usepackage{algorithm}
\usepackage{algorithmic}
\usepackage{amsthm}
\usepackage{caption}
\usepackage{wrapfig}
  
    \theoremstyle{plain}
    \newtheorem{theorem}{Theorem}[section]
    \newtheorem{lemma}{Lemma}[section]
    \newtheorem{definition}{Definition}[section]
    \newtheorem{proposition}{Proposition}[section]
    \newtheorem{setting}{Problem}
    \newtheorem{corollary}{Corollary}[section]

    \newtheoremstyle{TheoremNum}
        {\topsep}{\topsep}              
        {\itshape}                      
        {}                              
        {\bfseries}                     
        {.}                             
        { }                             
        {\thmname{#1}\thmnote{ \bfseries #3}}
    \theoremstyle{TheoremNum}
    \newtheorem{thmn}{Theorem}
    
    \newtheorem{lem}{Lemma}
    \newtheorem{prop}{Proposition}

\DeclareMathOperator*{\argmax}{argmax}

\newcommand\sH{\ensuremath{\mathcal{H}}}

\newcommand\sX{\ensuremath{\mathcal{X}}}

\newcommand\p[1]{\ensuremath{\left( #1 \right)}} 

\ifthenelse{\isundefined{\definition}}{\newtheorem{definition}{Definition}}{}
\ifthenelse{\isundefined{\assumption}}{\newtheorem{assumption}{Assumption}}{}
\ifthenelse{\isundefined{\hypothesis}}{\newtheorem{hypothesis}{Hypothesis}}{}
\ifthenelse{\isundefined{\proposition}}{\newtheorem{proposition}{Proposition}}{}
\ifthenelse{\isundefined{\theorem}}{\newtheorem{theorem}{Theorem}}{}
\ifthenelse{\isundefined{\lemma}}{\newtheorem{lemma}{Lemma}}{}
\ifthenelse{\isundefined{\corollary}}{\newtheorem{corollary}{Corollary}}{}
\ifthenelse{\isundefined{\alg}}{\newtheorem{alg}{Algorithm}}{}
\ifthenelse{\isundefined{\example}}{\newtheorem{example}{Example}}{}
\newcommand{\E}{\ensuremath{\mathbb{E}}} 

\newcommand\alphaSplitNeighborly{$\nicefrac{1}{\alpha}$-split-neighborly}

\begin{document}

%

%

\twocolumn[

\aistatstitle{Generalized Binary Search For Split-Neighborly Problems}

\aistatsauthor{ Stephen Mussmann \And Percy Liang }

\aistatsaddress{ Stanford University \And Stanford University } ]

\begin{abstract}
In sequential hypothesis testing, Generalized Binary Search (GBS) greedily chooses the test with the highest information gain at each step.
It is known that GBS obtains the gold standard query cost of $O(\log n)$ for problems satisfying the $k$-neighborly condition,
which requires any two tests to be connected by a sequence of tests where neighboring tests disagree on at most $k$ hypotheses.
In this paper, we introduce a weaker condition, split-neighborly,
which requires that for the set of hypotheses two neighbors disagree on, any subset is splittable by some test.
For four problems that are not $k$-neighborly for any constant $k$,
we prove that they are split-neighborly,
which allows us to obtain the optimal $O(\log n)$ worst-case query cost.

\end{abstract}

\section{Introduction}
Sequential hypothesis testing \citep{young1998sequential} aims to find the true hypothesis from a set of hypotheses by performing tests. Some examples are gathering observations to deduce the location of a hidden object or labeling data points to infer an underlying classifier. One commonly used algorithm is Generalized Binary Search (GBS), also known as the splitting algorithm, which greedily chooses the test that most evenly splits the hypothesis version space \citep{garey1974performance, nowak2008generalized}, or equivalently greedily chooses the test with the maximal information gain (for binary tests).  Greedy information gain is surprisingly effective in practice and has become the gold standard with a variety of applications, approximations, and extensions \citep{settles2012active,chu2005extensions,bellala2010extensions,karbasi2012comparison,zheng2012efficient,jedynak2012twenty,luo2013latent,maji2014active,sun2015active}.
We seek to explain this performance by providing a condition under which GBS attains a query cost of $O(\log n)$, the information-theoretic optimal query cost.

While there has been much work on establishing that GBS attains an average cost within a $\log n$ factor of the optimal algorithm \citep{guillory2009average, kosaraju1999optimal,dasgupta2004analysis,chakaravarthy2007decision,adler2008approximating,chakaravarthy2009approximating,gupta2010approximation}, establishing the asymptotically optimal query cost is a difficult and
understudied problem. The few previous works have required somewhat stringent
conditions. One such condition is ``sample-rich'' \citep{naghshvar2012noisy},
which states that every subset of the hypotheses has a test that returns true on exactly those hypotheses;
this requires an exponential number of tests.
Another line of work
\citep{nowak2009noisy,nowak2011geometry} introduced the more lenient
\emph{$k$-neighborly} condition, which requires that every two tests be connected by a sequence of tests
where neighboring tests disagree on at most $k$ hypotheses.
As we will show in this paper, many problems of a discrete nature do not satisfy this condition.

In this paper, building on the $k$-neighborly condition,
we introduce a new, weaker condition called
\emph{\alphaSplitNeighborly}, which requires that if neighboring tests
disagree on a set of hypotheses $V$, then there exists a test that splits off
an $\alpha$ fraction of $V$ (note that $|V|$ could be quite large, whereas $k$-neighborly requires $|V| \le k$).
We prove that four natural problems satisfy the \alphaSplitNeighborly\ condition for a constant $\alpha$:
pool-based linear classifiers, learning monotonic CNF formulas, discrete object localization, and discrete binary classification.
Furthermore,
we prove that the value of $k$ in the $k$-neighborly analysis is at least $\sqrt{n}/2$ for all four problems,
which yields nearly vacuous bounds.
In summary, by using \alphaSplitNeighborly, we show that Generalized Binary Search achieves an asymptotically
optimal query cost of $O(\log n)$ in settings where the previous $k$-neighborly analysis tools fail.

\subsection{Notation}
\label{sec:notation}
In all cases, we use $\log$ to denote $\log_2$. For a set $S$ and function $f$, we define ${\mathbb{E}_{s \in S}[f(s)] = \frac{\sum_{s \in S} f(s)}{|S|}}$. Similarly, for a condition $C$, we define ${\Pr_{s \in S}[C(s)] = \frac{\sum_{s \in S} \mathbf{1}[C(s)]}{|S|}}$.
\section{Problem statement}
Consider a set of $n$ hypotheses $\mathcal{H}$ and tests $\mathcal{X}$,
where each $h \in \mathcal{H}$ is a mapping from $\mathcal{X}$ to $\{0, 1\}$. We assume that the hypotheses $\sH$ are identifiable, meaning that any two hypotheses yield different results on at least one test.
Assume there is a fixed but unknown hypothesis $h^* \in \mathcal{H}$ that we wish to identify.
An active querying algorithm performs a sequence of tests;
on each iteration, it uses a method $F$ to select a test $x_t$ based on the results of previous tests
and receives $y_t = h^*(x_t)$ (see Algorithm~\ref{alg:generic}).
We evaluate $F$ based on the \emph{worst-case} number of queries.

\begin{figure}[t!]
  \begin{minipage}[t]{.98\linewidth}

    \begin{algorithm}[H]
      \caption{Active querying algorithm template} 
      \label{alg:generic}
      \begin{algorithmic}
        \STATE {\bfseries Input:} $\mathcal{H}$, $\mathcal{X}$, oracle access to $h^*$, method $F$
        \STATE $V = \mathcal{H}$
        \FOR{$t=0,1,...$}
        \STATE $x_t \leftarrow F( \{(x_i, y_i)\}_{i=1}^{t-1} )$
        \STATE Query $x_t$ and obtain $y_t = h^*(x_t)$
        \STATE Update $V \leftarrow \{h \in V: h(x_t) = y_t\}$
        \IF{ $|V|=1$ }
        \STATE {\bfseries return} $h \in V$
        \ENDIF
        \ENDFOR
      \end{algorithmic}
    \end{algorithm}

  \end{minipage}
  \hfill
  \begin{minipage}[t]{.98\linewidth}

    \begin{algorithm}[H]
      \caption{Generalized Binary Search ($F$)}
      \label{alg:GBS}
      \begin{algorithmic}
        \STATE {\bfseries Input:} $\mathcal{H}$, $\mathcal{X}$, Previous test results $\{(x_i, y_i)\}_{i=1}^{t-1}$
   \STATE $V = \{h \in \mathcal{H}: h(x_i) = y_i, 1 \leq i \leq t-1\}$
   \STATE $x_t \leftarrow \text{argmin}_{x \in \mathcal{X}} |\mathbb{E}_{h \in V}[h(x)] - 1/2|$
   \STATE {\bfseries return} $x_t$
\end{algorithmic}
\end{algorithm}
\end{minipage}
\end{figure}

\begin{figure}
\centering
  \includegraphics[width=.98\linewidth]{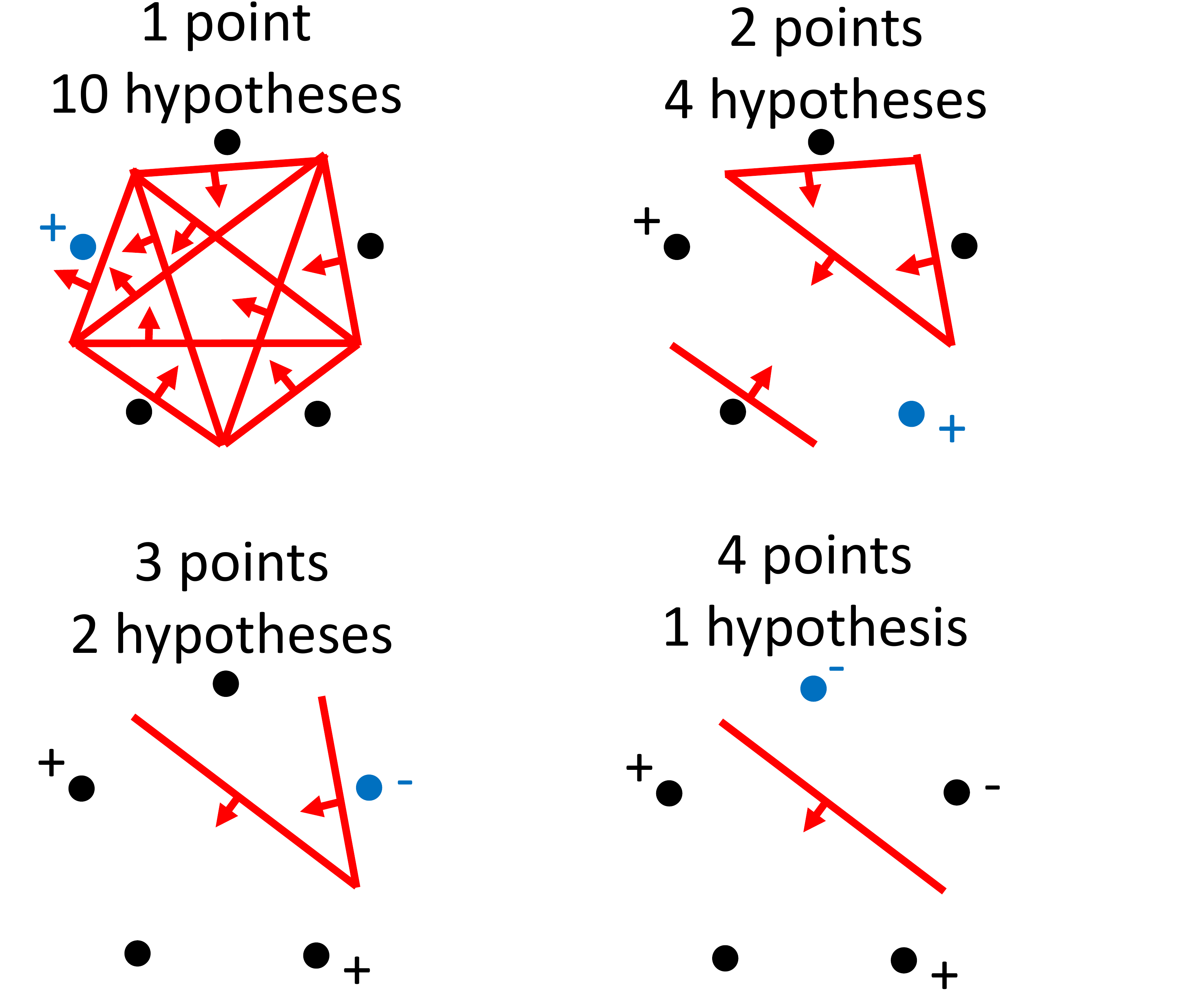}
  \caption{
    Problem of identifying a linear classifier
in a pool-based active learning setting.
  Each linear classifier is represented by a red line with an arrow pointing towards the positive class.
  Each round, we select a new test (blue point), after four rounds,
  we have identified the true classifier.}
    \label{fig:classifier-example}
\end{figure}

As an illustrative example, let the set of hypotheses $\mathcal{H}$ be linear classifiers separating $5$ data points at the vertices of a regular pentagon.
(see Figure~\ref{fig:classifier-example}). In this case, $|\mathcal{H}|=20$ (only including hypotheses with both $+$ and $-$ labels). The tests $\mathcal{X}$ are data points, and the test output is an indicator for $h^*$ classifying that point as $+$.
Figure \ref{fig:classifier-example} shows the queries generated by GBS.
We see that the size of the version space $|V|$ decreases exponentially,
the hallmark of a $O(\log n)$ worst-case query cost.
Later, we will prove that GBS indeed attains $O(\log n)$ worst-case query cost for this
this problem class of linear classifiers on the vertices of convex polygons.

Though we focus on the noiseless and well-specified setting,
both conditions can be relaxed:
\citet{kaariainen2006active} reduces the non-persistent noisy setting (we can repeatedly query any test) to the noiseless setting,
and \citet{nowak2011geometry} adapts
the GBS algorithm to the mis-specified setting (see Section~\ref{sec:discussion} for more details).

Generalized Binary Search (also known as the ``splitting algorithm'' and ``maximal shrinkage'') is a well-studied method \citep{garey1974performance, nowak2011geometry,dasgupta2004analysis}. GBS maintains the set $V \subseteq \sH$ of hypotheses consistent with test results thus far, and at each step, it chooses a test that splits the elements of $V$ as evenly as possible. See Algorithm \ref{alg:GBS} for the pseudocode. The optimal worst-case number of queries is $\Omega(\log n)$ so if GBS attains $O(\log n)$ for a problem, it is asymptotically optimal.

\section{General analysis}
\subsection{Splits}
Intuitively, GBS works well when it can find tests that split the hypothesis space into roughly equal parts.
The \textit{split} induced by test $x$ is a partition of the hypotheses into $\{h \in \sH: h(x)=0\}$ and $\{h \in \sH: h(x)=1\}$.
Define the \textit{split constant} of a test $x$ for a set of hypotheses $V$  as $\min_{y \in \{0,1\}} \Pr_{h \in V}[h(x)=y]$,
the fraction of hypotheses in the smaller partition.
Note that large split constants are preferred, and $1/2$ is the maximum split constant.
As we will see, both the $k$-neighborly condition \citep{nowak2011geometry} and
our new \alphaSplitNeighborly\ condition imply that for any version space $V$, there is a test with a large split constant.

\subsection{Earlier work: k-neighborly and coherence parameter}
\begin{figure}
  \centering
  \includegraphics[width=.98\linewidth]{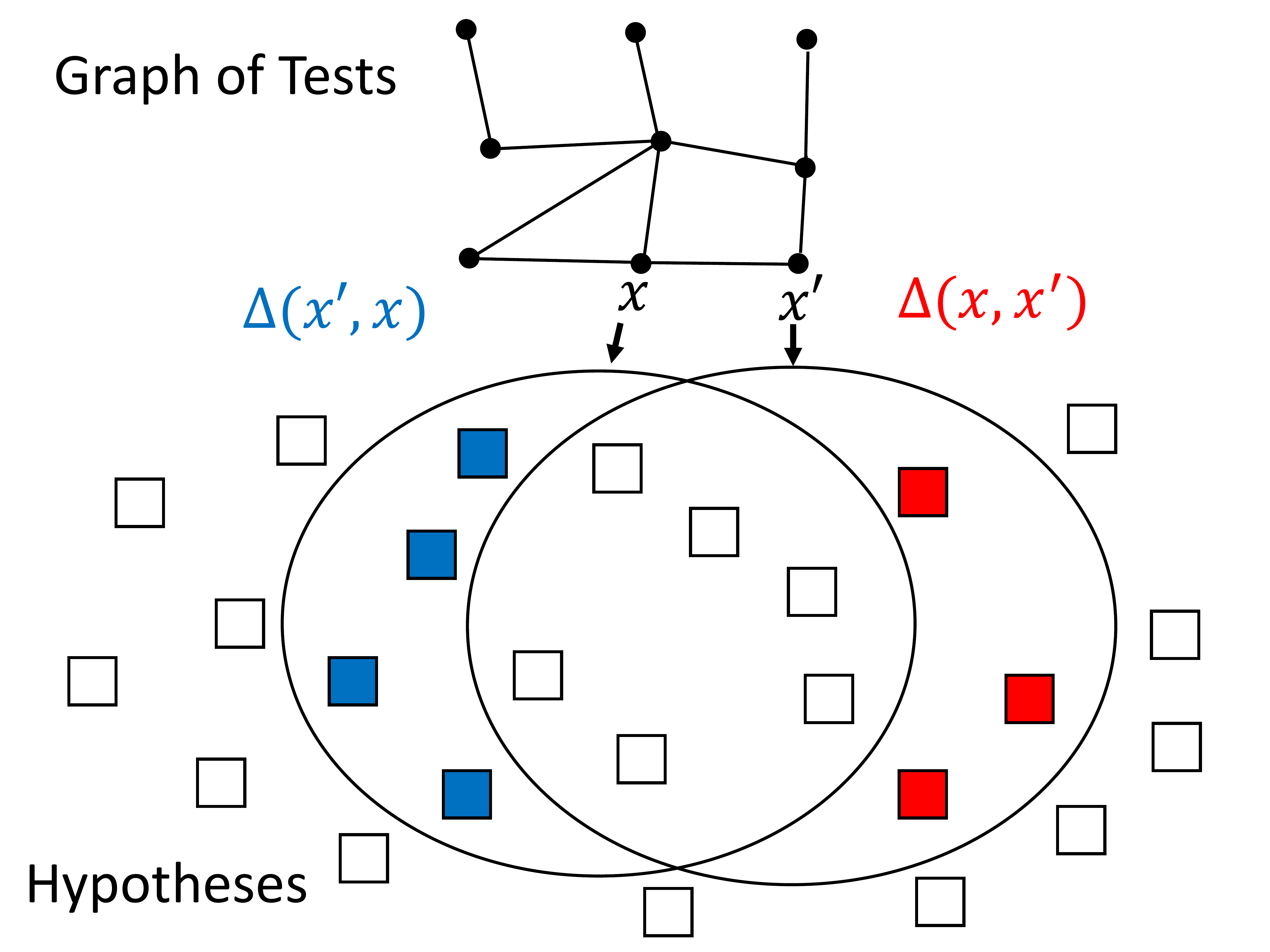}
\hfill
  \caption{A test graph on the top and the action of two neighboring tests $x,x'$ on the hypotheses on the bottom.
  The hypotheses $h$ are represented by rectangles and the tests by circles that contain the hypotheses for which the test returns $1$.
  For a $k$-neighborly edge to exist between two tests, the corresponding $\Delta(x,x')$ and $\Delta(x',x)$ must have cardinality $|\Delta(x,x') \cup \Delta(x',x)| \leq k$. If the resulting test graph is connected, we say the problem is $k$-neighborly.}
  \label{k-neighborly}
\end{figure}

Tests often have a similarity structure.
As an example, for the hypothesis class of linear classifiers, nearby input points (tests) yield the same result for most hypotheses.
We therefore construct a similarity graph over tests,
which will provide a useful analysis tool that allows us to only reason locally on the graph.
\citet{nowak2011geometry} defines two tests to be similar if they disagree on at most $k$ hypotheses.
$k$-neighborly is the condition that such a similarity graph is connected.
\begin{definition}[$k$-neighborly]
For any two tests, $x$ and $x'$, define $\Delta(x,x') = \{h \in \mathcal{H}: h(x)=0 \wedge h(x')=1\}$.
Let the test graph contain undirected edges $(x,x')$ for which $|\Delta(x,x') \cup \Delta(x',x)| \le k$.
A problem instance is $k$-neighborly if the test graph is connected.
\end{definition}
See Figure \ref{k-neighborly} for an illustration of the $k$-neighborly condition. Intuitively, the $k$-neighborly condition ensures that between any two tests, we can find a path where each pair of neighbors in the path are very similar.
\citet{nowak2011geometry} also defines the coherence parameter,
which ensures an algorithm can easily find tests that both return $0$ and $1$ by choosing tests randomly.\footnote{
  Our definition is a simple linear transformation of the definition in \citet{nowak2011geometry} to account for notational differences.}
\begin{definition}[Coherence parameter]
  The coherence parameter is the largest $c$ such that
  $$\forall h \in \mathcal{H} : \E_{x \sim P}[h(x)] \in [c, 1-c]$$
  for some probability distribution $P$ over tests.
\end{definition}
This is a concept that will be used with our condition, \alphaSplitNeighborly, as well. From these two definitions, \citet{nowak2011geometry} showed the following result:

\begin{theorem}[Nowak, 2011]
If a problem has a coherence parameter $c$ and is $k$-neighborly, then the worst-case cost of GBS is $\frac{1}{-\log(\lambda)} \log(n)$ queries, where $\lambda = 1 - \min(c, \frac{1}{k+2})$.
\end{theorem}

For large enough $c$, the $k$-neighborly analysis yields worst-case query complexity of $O(k \log(n))$. Later, we show several examples where $k=\Omega(\sqrt{n})$, yielding the $k$-neighborly analysis very loose.
\subsection{Split-neighborly}
The $k$-neighborly condition is a rather strong condition since it requires tests that disagree on only $k$ hypotheses. While this sometimes holds for problems with a continuous structure, such as linear classifiers or continuous object localization, it is often not satisfied for problems with a discrete nature.
Later, in Section \ref{sec:applications}, we show a variety of problems of a discrete nature where $k$ is at least $\sqrt{n}/2$. Motivated by these discrete problems, we will now introduce a weaker condition which we call \alphaSplitNeighborly,
the main contribution of this paper.
In \alphaSplitNeighborly,
two tests are not only connected if there is a small number of hypotheses on which the
tests differ, but also if any subset of the hypotheses that they differ on can be split evenly (with a split constant of at least $\alpha$) by
some test.

\begin{definition}[\alphaSplitNeighborly]
  Let $\alpha \in (0,\frac{1}{2}]$ \footnote{As a special case, we say a problem is $1$-split-neighborly if the graph generated by connecting nodes where $|\Delta(x,x')|\leq 1$ is strongly connected.}. For any two tests, $x$ and $x'$, define $\Delta(x,x') = \{h \in \mathcal{H}: h(x)=0 \wedge h(x') = 1\}$.
Define a directed test graph to have a directed edge $(x,x')$ if for any $V \subseteq \Delta(x,x')$, $|V| \leq 1$ or there exists a test $x \in \sX$ such that 
$$\mathbb{E}_{h \in V}[h(x)] \in [\alpha, 1-\alpha]$$
A problem is \alphaSplitNeighborly\ if the test graph is strongly connected.
\end{definition}

\begin{figure}
\centering
  \includegraphics[width=.98\linewidth]{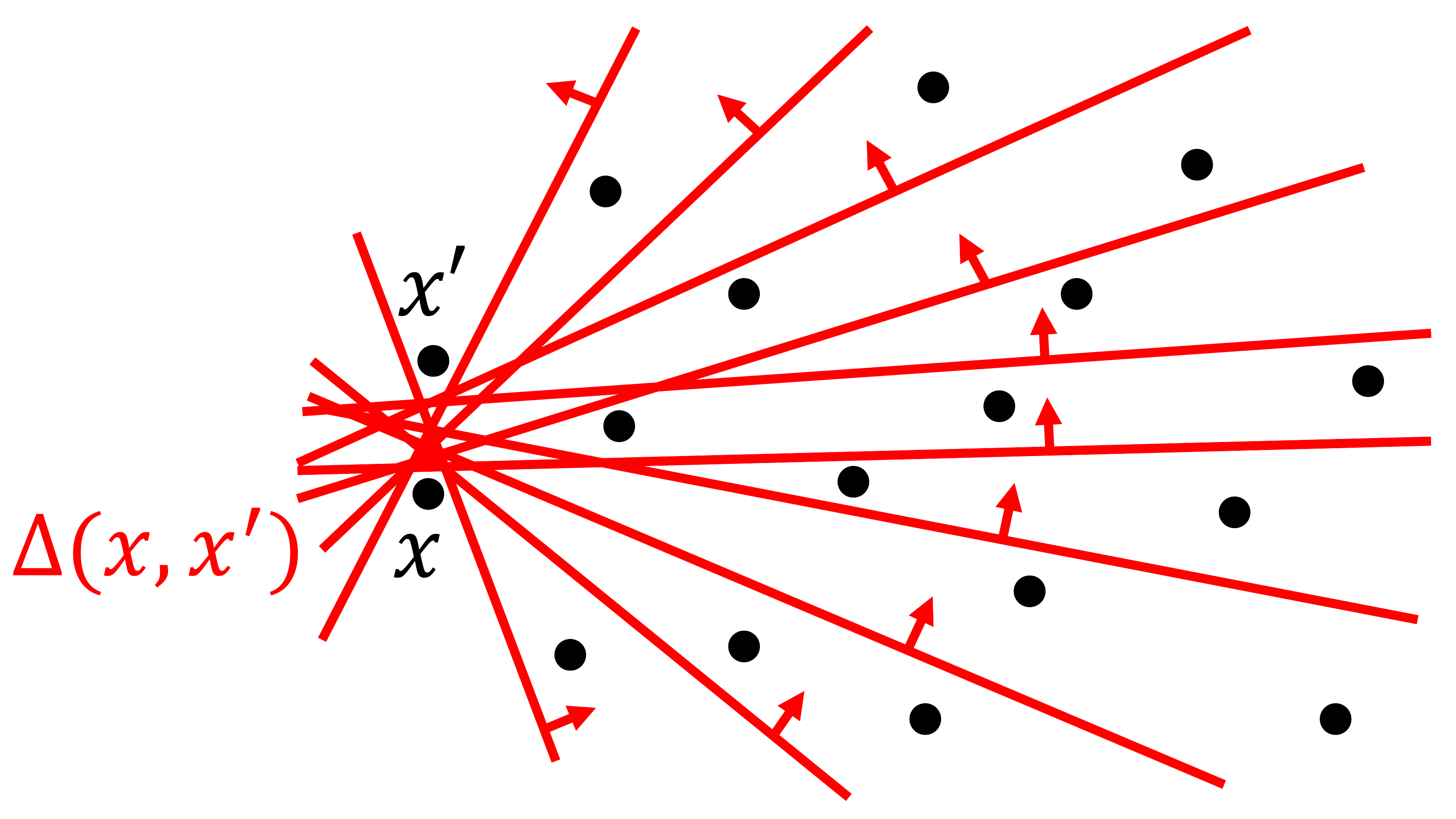}
  \caption{For the problem of identifying linear classifiers in a pool-based active learning setting,
  an example of two tests that are not connected in the $k$-neighborly graph for small $k$ but are connected in the \alphaSplitNeighborly\ graph for small $\nicefrac{1}{\alpha}$.
  While the size of $\Delta(x,x')$ is large, we can still split any subset of $\Delta(x,x')$ because of the other points in the pool.}
    \label{fig:split-neighborly}
\end{figure}

Although it can be more involved to show an edge between tests in the sense of \alphaSplitNeighborly\ rather than $k$-neighborly, it is a more general condition which makes the similarity graph more connected (see Figure \ref{fig:split-neighborly} for an example).

The coherence, $k$-neighborly, and \alphaSplitNeighborly\ conditions are preserved when we restrict the hypotheses: create a problem with same tests $\sX' = \sX$ but with $\sH' \subseteq \sH$. This is because the conditions all are statements involving a universal quantification over the hypotheses, or subsets thereof.

Furthermore, and most importantly, constant coherence and the split-neighborly condition imply that GBS has $O(\log n)$ query cost. First, we prove the following lemma, showing that constant coherence and \alphaSplitNeighborly\ imply that any subset of $\sH$ has a test with a good split constant.

\begin{lemma}
\label{split-neighborly-lemma}
If a problem is \alphaSplitNeighborly\ and has a coherence parameter of $c$, then for any $V \subseteq \sH$, $|V| \leq 1$ or there exists a test $x \in \sX$ such that 
$$\mathbb{E}_{h \in V}[h(x)] \in [\beta, 1-\beta]$$
  where the split constant is
  $$\beta = \min\p{c, \frac{1}{\nicefrac{1}{\alpha}+2}}.$$
\end{lemma}
Note that for large $c$ and small $\alpha$, $\beta \approx \alpha$. See appendix for the full proof; we only give a sketch here. Intuitively, the coherence parameter ensures that there is a good split of $V$ or there is both a test that mostly yields $0$ and a test that mostly yields $1$ (for hypotheses $V$).
If we examine a path of tests $x$ between the two tests,
either $\mathbb{E}_{h \in V}[h(x)]$ varies smoothly from close to $0$ to close to $1$,
in which case there is a good split, or there is a large jump in the split constant between two neighboring tests $x$ and $x'$,
which implies that $|V \cap \Delta(x, x')|/|V|$ is large.
Finally, from the definition of \alphaSplitNeighborly, we can find a test to have a $\beta$ split constant of $V \cap \Delta(x, x')$. In summary, the split-neighborly condition and coherence condition allow us to conclude that for any subset of the hypotheses, there is a test with a $\beta$ split constant.

From this lemma, we get the following theorem.

\begin{theorem}
\label{split-neighborly}
If a problem is \alphaSplitNeighborly\ and has a coherence parameter of $c$,
  then GBS has a worst case query cost of at most $\frac{\log n}{-\log (1-\beta)}$,
  where $$\beta = \min\p{c, \frac{1}{\nicefrac{1}{\alpha}+2}}.$$
\end{theorem}

From Lemma \ref{split-neighborly-lemma}, it is clear that after $m$ queries, we have at most $n (1 - \beta)^m$ hypotheses left.
Thus, the worst-case query cost (to reach one hypothesis) is 
$\frac{\log n}{-\log(1-\beta)}$. The precise proof is in the appendix.

Similarly to the $k$-neighborly condition, for large enough coherence, the worst-case query cost is $O(\frac{1}{\alpha} \log n)$. Thus, for constant $\alpha$, we get $O(\log n)$ worst-case query cost, but for $\alpha \rightarrow 0$, we do not.

\begin{figure}
\centering
  \includegraphics[width=.6\linewidth]{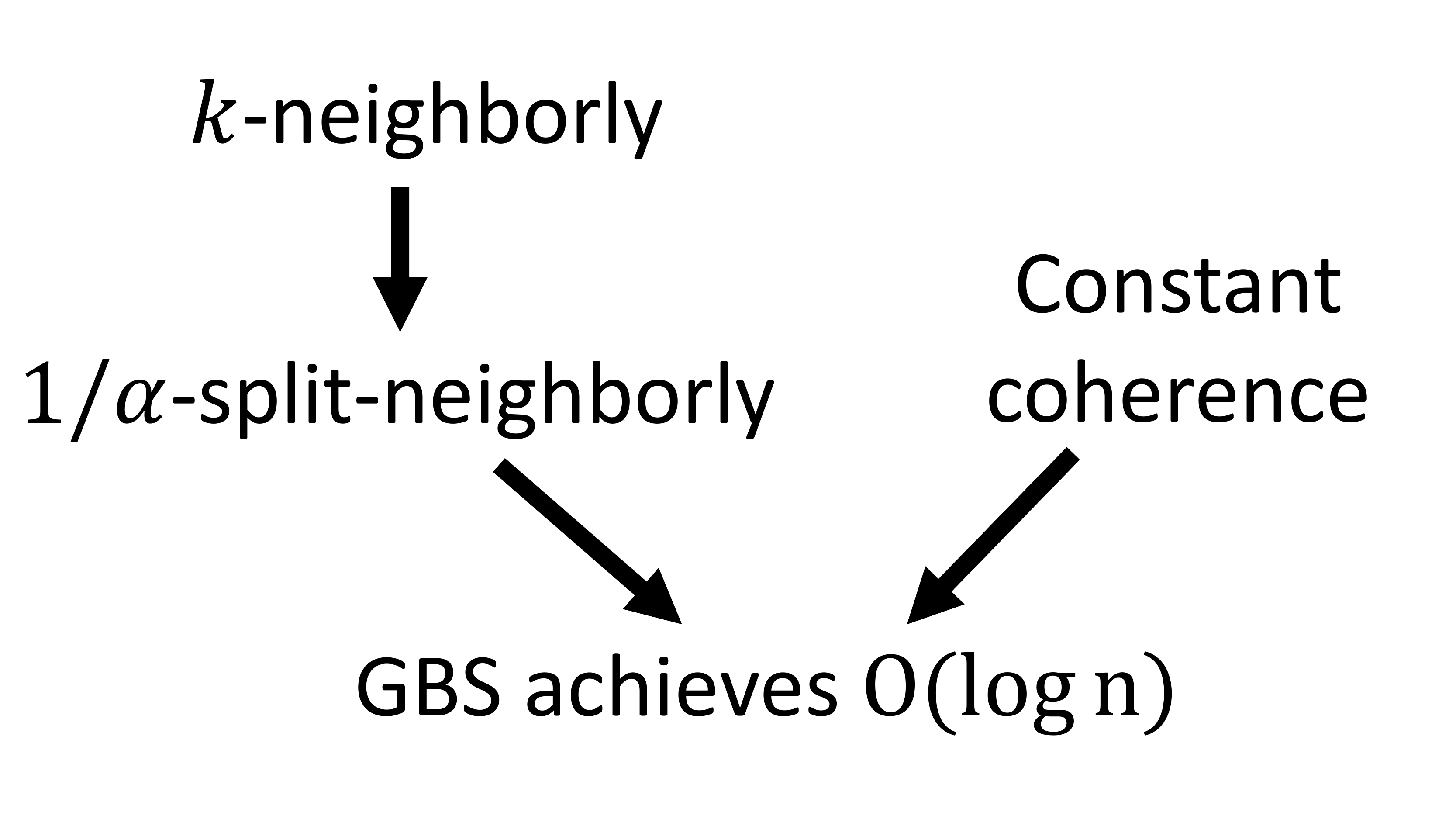}
  \caption{Relationship between the different conditions, where arrows represent logical implication.}
  \label{fig:diagram}
\end{figure}

In fact, $k$-neighborly implies $k$-split-neighborly (\alphaSplitNeighborly, $\alpha = 1/k$). Thus, our split-neighborly condition is a generalization of $k$-neighborly, and comparison between our theorems shows our condition is strictly more powerful than the $k$-neighborly condition. See Figure \ref{fig:diagram} for a diagram.

\begin{proposition}
\label{k-neighborly-to-split-neighborly}
If a problem is $k$-neighborly, then it is $k$-split-neighborly.
\end{proposition}
\begin{proof}
In the case that $k=1$, $|\Delta(x,x')|=1$ so $|V|\leq 1$ so the problem is $1$-split-neighborly. Note that any set of hypotheses must have a test that distinguishes at least one of the hypotheses (otherwise the hypotheses are the same). If two points $x$ and $x'$ in the $k$-neighborly graph have an edge between them, then $|\Delta(x,x') \cup \Delta(x',x)| \leq k$, which implies $|V| \leq |\Delta(x,x')| \leq k$, and thus either $|V| \leq 1$ or there is a test with a $1/k$ split constant and thus there is an edge from $x$ to $x'$ in the $k$-split-neighborly graph. Similarly, there is an edge from $x'$ to $x$, the $k$-split-neighborly graph is strongly connected, and the problem is $k$-split-neighborly.
\end{proof}

\section{Application of analysis}
\label{sec:applications}

In this work, we establish the \alphaSplitNeighborly\ condition for four problems: two-dimensional linear classifiers on the vertices of a convex polygon, learning monotonic disjunctions and CNF formulas, discrete object localization (under two different conditions), and discrete linear classifiers. We show that GBS achieves $O(\log n)$ cost on these problems under conditions on $\sH$ by showing that the problems are \alphaSplitNeighborly\ and have constant coherence. Further, we show the inadequacy of the $k$-neighborly analysis for each of these problems. 

All of the proofs have a similar structure for proving \alphaSplitNeighborly. First, fix a subset of hypotheses $V \subseteq \Delta(x,x')$.
Then, by assuming there is no test with a good split constant $\alpha$, we can leverage the structure of the problems to conclude that the size of $V$ is small. Since any two hypotheses disagree on at least one test (identifiability of hypotheses), we can always split off one of the hypotheses for a split constant of $1/|V|$ which is a good split if $|V| \leq \nicefrac{1}{\alpha}$.

Since there is no test with a split constant $\alpha$, any test either yields $1$ on the vast majority of hypotheses in $V$ or yields $0$ on the vast majority of hypotheses in $V$. Thus, we partition the tests into two sets,
$$\sX^+ = \{x \in \sX: \Pr_{h \in V}[h(x)=1] > 1-\alpha\}$$
$$\sX^- = \{x \in \sX: \Pr_{h \in V}[h(x)=1] < \alpha\} = \sX - \sX^+$$
Several of the arguments will leverage the structure of this partition and use union bound to show that the probability of a single hypothesis is high, and thus $V$ is small.

\subsection{Two-dimensional linear classifiers with convex polygon data pool}

Suppose we have a pool of unlabeled data points and our set of hypotheses is linear classifiers in the transductive setting (we group all hypotheses with the same output on all unlabeled data points together). We examine the case of two dimensions. 

\begin{setting}[Linear classifiers on convex polygon data pool]
  Let $\sX$ be a set of $m$ points $x \in \mathbb{R}^2$ such that the points are the vertices of a convex polygon. Let $\sH$ be equivalence classes of linear classifiers that have the same output on $\sX$ and such that
$$\frac{\sum_{x \in \sX} h(x)}{|\sX|} \in \left[\frac{1}{4},\frac{3}{4}\right].$$
\end{setting}
This last constraint restricts the classifiers to those with balanced labels.
This ensures a good coherence parameter; otherwise, no algorithm can perform better than $\Theta(m) = \Theta(\sqrt{n})$.

\begin{figure}[t!]
\centering
  \includegraphics[width=.9\linewidth]{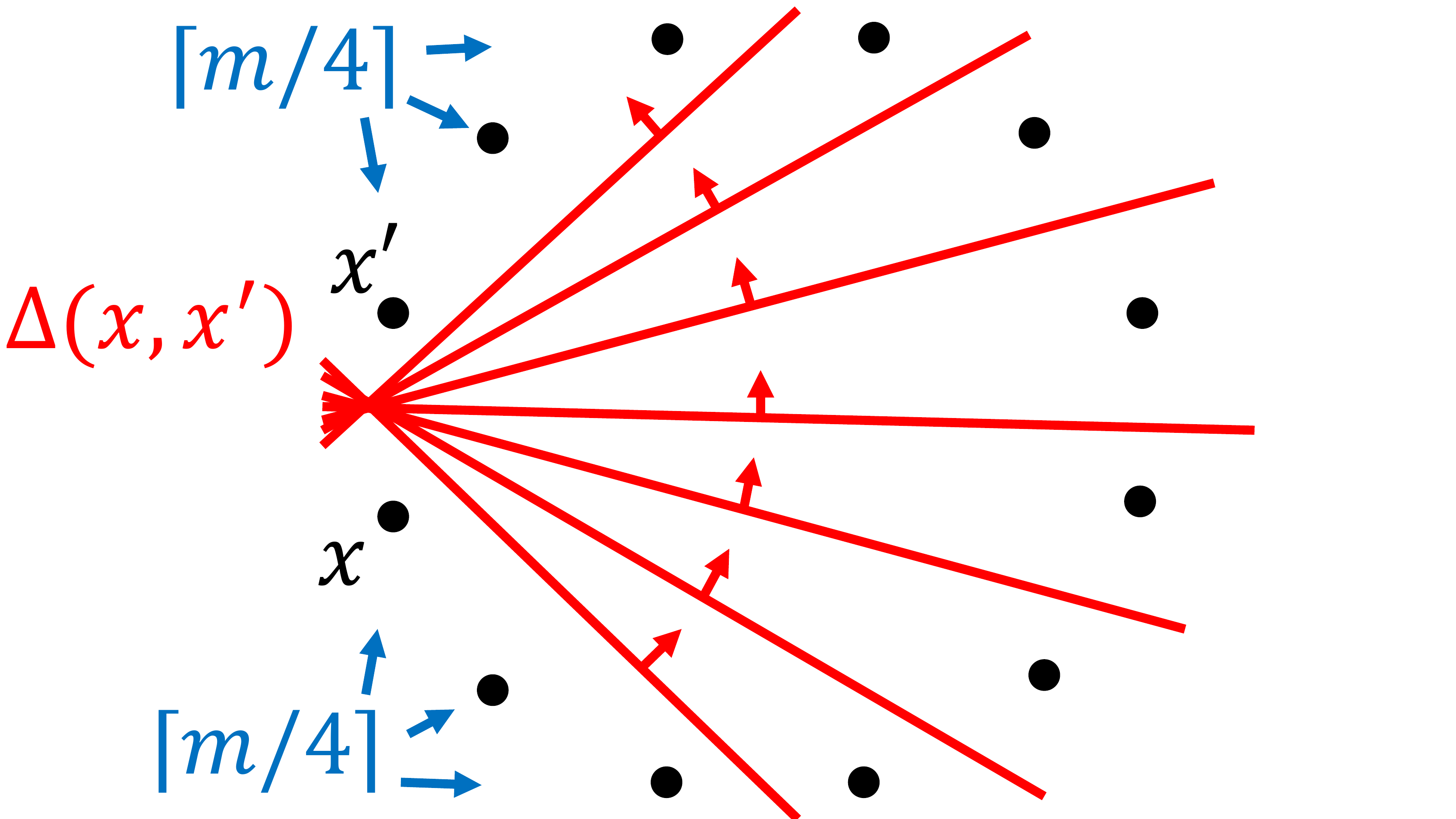}
  \caption{An illustration of $\Delta(x,x')$ for linear classifiers on a data pool forming the vertices of a convex polygon. Note that we can split any subset of $\Delta(x,x')$ with a split constant of at least $1/3$ because the tests are interleaved in a sequence with the hypotheses.}
  \label{fig:convex-data-pool}
\end{figure}

We will now show that the $k$-neighborly analysis for this problem is poor.
See Figure \ref{fig:convex-data-pool} for a diagram. For adjacent points $x$ and $x'$, $|\Delta(x,x')| = m - 2\lceil m/4 \rceil + 1$. Further note that $n = |\sH| = m(m - 2\lceil m/4 \rceil + 1)$ \footnote{We have 2 linear classifiers for each line, but we are double counting.}. Thus, $|\Delta(x,x')| \geq \frac{\sqrt{n}}{2}$ (for $m\geq 4$) and so the $k$ for the $k$-neighborly analysis is at least $\frac{\sqrt{n}}{2}$.

However, it is clear from Figure \ref{fig:convex-data-pool} that $\Delta(x,x')$ is a sequence of hypotheses with tests interleaved. Thus, we can split $\Delta(x,x')$ with at least a split constant of $1/3$ and to get the following proposition,

\begin{proposition}
  The problem of learning a linear classifier on a convex polygon data pool is $3$-split-neighborly.
\end{proposition}

Note that because of the constraint that the minority label is at least $1/4$, the coherence parameter is at least $c=1/4$. Thus, the worst case query complexity is at most $\frac{\log n}{-\log(1 - 1/5)} \leq 3.2 \log n$.

\subsection{Monotonic CNF formulas}

In this section, we examine the problem of learning monotonic CNF formulas from function evaluations. To begin, we study the case of a single disjunction, such as the following,
$$x_4 \vee x_7 \vee x_9$$
\begin{setting}[Disjunction]
\label{ps:disjunction}
Let the elements of $\mathcal{H}$ be a disjunction over $d$ variables without any negations where the disjunction has at most $m$ variables. Let $\mathcal{X}$ be the set of length $d$ bit assignments.
\end{setting}

First, note that $h(0^d)=0$ and $h(1^d)=1$ for all $h \in \sH$ and thus the coherence parameter is $c=1/2$. The $k$-neighborly analysis is lacking for this problem. Note, $|\sH| = \sum_{i=1}^m \binom{d}{i}$. However, the bit string $0^d \in \mathcal{X}$ disagrees with all other $x \in \sX$ for $\sum_{i=1}^m \binom{d-1}{i-1}$ hypotheses. So for $m \geq 2, d \geq 2m$, then $k \geq \sqrt{n}$. See the appendix for details. On the other hand, our split neighborly analysis achieves the optimal rate in the case where $m$ is constant and $d$ goes to infinity.

\begin{figure}[t!]
\centering
  \includegraphics[width=\columnwidth]{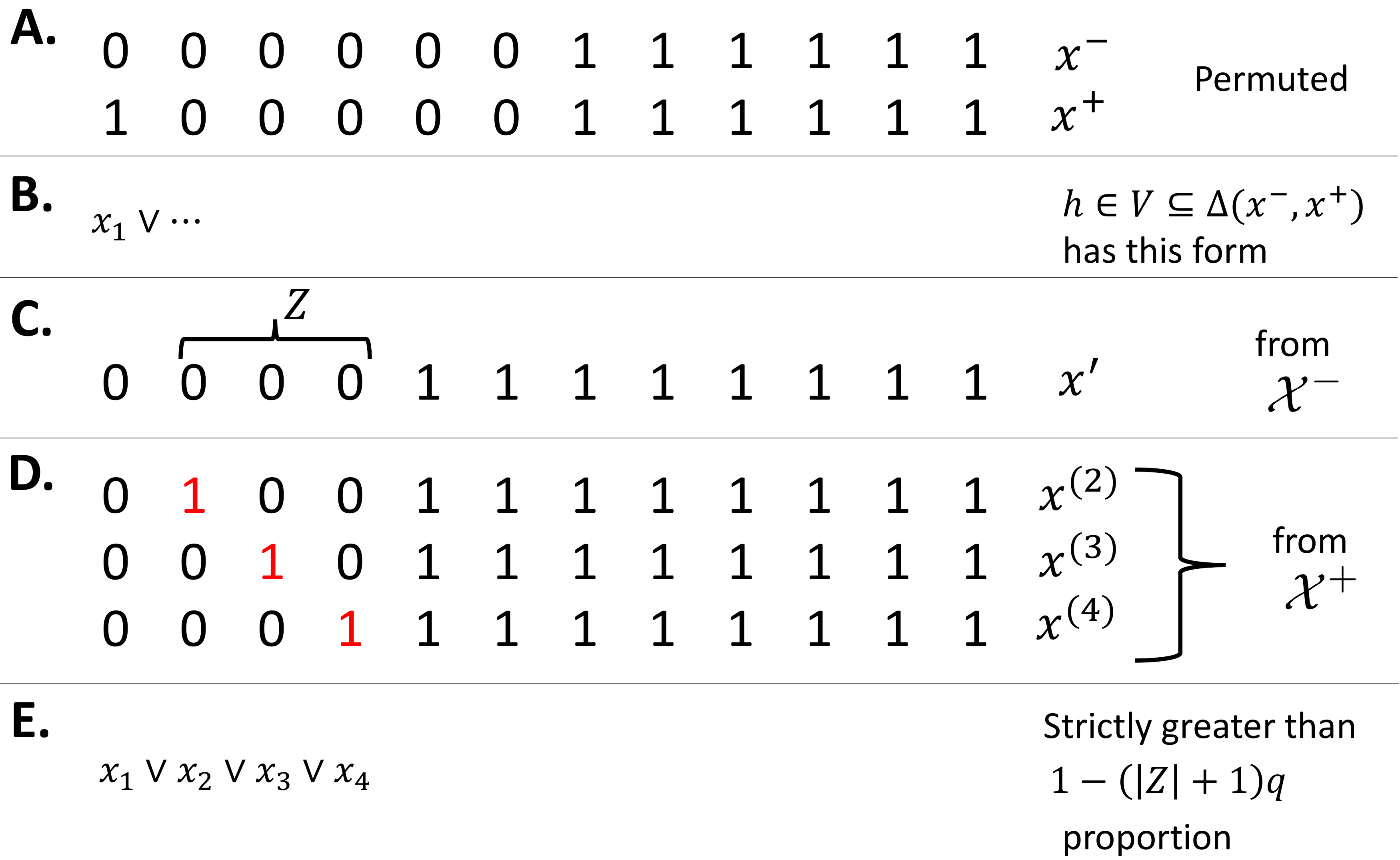}
  \caption{A proof illustration for the disjunction problem being split-neighborly.}
  \label{fig:disjunction_proof}
\end{figure}

\begin{theorem}
The single disjunction problem is ${(m+1)}$-split-neighborly.
\end{theorem}
\begin{proof}
A graphic for the proof is shown in Figure \ref{fig:disjunction_proof}.

We will show that there are edges between tests that differ by just one bit. This will suffice since such a graph is strongly connected. In particular, we show that the test graph has a bidirectional edge from $x$ to $x'$ if $||x - x'||_1 = 1$.

Let $x^+$ be the value of $x$ or $x'$ with more $1$'s (and let $x^-$ be the other one). Note that from monotonicity, $|\Delta(x^+,x^-)| = 0$ so there is a directed edge from $x^+$ to $x^-$.

For the other direction, fix a subset $V \subseteq \Delta(x^-,x^+)$. Without loss of generality, let $x^+$ and $x^-$ differ in the first coordinate so $x^+_1 = 1$ and $x^-_1=0$ and $\forall i>1: x^+_i = x^-_i$. See row A of Figure \ref{fig:disjunction_proof}. Because $V \subseteq \Delta(x^-,x^+)$, all hypotheses in $V$ include $x_1$ in the disjunction. See row B of Figure \ref{fig:disjunction_proof}.

For ease of notation, let $q=1/(m+1)$. We will proceed by showing that if there is no test with a good split, $\mathbb{E}_{h \in V}[h(x)] \in [q, 1 - q]$, then $|V| \leq m+1$. Then, either $|V| \leq 1$ or there is a test with split constant at least $1/(m+1)$ and the proof is complete.

Now, if there are no tests with a good split, each test must either yield $1$ or $0$ for the vast majority of hypotheses in $V$. Thus, we can define the following two sets.
$$\sX^+ = \{x \in \sX: \Pr_{h \in V}[h(x)=1] > 1-q\}$$
$$\sX^- = \{x \in \sX: \Pr_{h \in V}[h(x)=1] < q\} = \sX - \sX^+$$
Let $x'$ be the the element of $\sX^-$ with the fewest $0$'s. Since $x' \in \sX^-$, $x'_1=0$. Let $Z$ be the other indices of the $0$'s. If $|Z|=0$, then $|\Delta(x^-,x^+)|=|\{x_1\}|=1$ so $|V|\leq 1$ and we are done. Define $\{x^{(j)}\}_{j \in Z}$ to be the test resulting from $x'$ and changing the $j^{th}$ bit to a $1$. By the minimal definition of $x'$, $\forall j \in Z: x^{(j)} \in \sX^+$. See rows C and D of Figure \ref{fig:disjunction_proof}.

We now derive a useful equation. Note that for any subset $Z' \subseteq Z$, from the definition of $\sX^+$ and $\sX^-$ and union bound, $\Pr_{h \in V}[h(x')=0 \wedge \forall j \in Z': h(x^{(j)})=1] > 1 - (|Z'|+1)q$. From the property of disjunctions, this implies $\Pr_{h \in V}[h \text{ has variables at } Z' \cup \{1\}] > 1 - (|Z'|+1)q$. See row E of Figure \ref{fig:disjunction_proof}.

If $|Z| \geq m$, then this means that we can choose a subset $Z'$ of size $m$. $\Pr_{h \in V}[h \text{ includes } m+1 \text{ variables} ] > 1 - (m+1)q = 0$. This means there is a non-zero probability of a hypothesis with $m+1$ variables which is impossible, since our disjunctions don't have more than $m$ variables. So $|Z| \leq m-1$.

We are nearly done. Note that the left side of the useful equation is exactly $1/|V|$. Therefore, $1/|V| > 1 - (|Z|+1)q \geq 1-mq \geq 1/(m+1)$. Rearranging, we find that $|V| < m+1$ and we are done.
\end{proof}

We now examine the more general monotonic CNF problem from function evaluations. An example of such a monotonic formula is:
$$(x_1 \vee x_4 \vee x_5) \wedge (x_2 \vee x_7 \vee x_8).$$
\begin{setting}[Conjunction of disjunctions]
\label{ps:conj_disjunction}
  Let $\mathcal{H}$ be a conjunction of $\ell$ $m$-disjunctions over $d$ variables without any negations, and where each variable does not appear in multiple disjunctions. Let $\mathcal{X}$ be the set of length $d$ bit assignments.
\end{setting}

Note that there is an isomorphism between conjunctions of disjunctions and disjunctions of conjunctions by flipping the test bits and the result bit. 

Additionally, for a general setting shown in the appendix, $k \geq \sqrt{n}$ which renders the $k$-neighborly analysis very poor.

However, the split-neighborly analysis suffices,

\begin{theorem}
\label{thm:cnf}
The conjunction of disjunctions problem is $(m+1+3(l-1))$-split-neighborly.
\end{theorem}

The proof is in the appendix. Note that the value of $\nicefrac{1}{\alpha}$ does not depend on the number of variables, so GBS is efficient even for very large $d$ when $m,l$ are constant.


\subsection{Object localization in $\mathbb{Z}^d$}

Consider the problem of object localization \citep{chen2015submodular} where we try to locate an object based on spatial queries. In this work, we wish to find the location $z$ of an object in space or in an image by asking queries of the form ``Is $z$ close to point $x$?''. We can discretize the space into the grid of integers and define ``closeness'' by as whether $z - x$ is in some set $S$ (e.g., an $\ell_p$ ball).

In this way, hypotheses and tests are both indexed by vectors of integers. For concreteness, for an $\ell_p$ norm ball, a test $x$ returns the result of $\|x-z\|_p \leq \ell$. 

\begin{setting}[Object localization]
\label{ps:lp_norm}
  Fix a set $S \subseteq \mathbb{Z}^d$ representing the sensing field
\begin{align*}
\mathcal{H} &\subseteq \{h_z\}_{z \in \mathbb{Z}^d} &
\mathcal{X} &= \mathbb{Z}^d &
h_z(x) &= \mathbf{1}[ x-z \in S]
\end{align*}
\end{setting}

Note that there are infinitely many integer vectors, but we can take a bounded region as the hypothesis space. Note that if the bounded region of the hypothesis space is too large, the coherence parameter would be very small,
and greedy would not perform well (and any algorithm for that matter, since the algorithm would have to do a linear search).
One way to make the coherence parameter $c=1/2$ is to choose a $x^*$ and ensure that $\mathcal{H} \subseteq \{h_z : z - x^* \in S\}$.

For cases where $S$ is an axis-symmetric box, or equivalently where we use a weighted $\ell_\infty$ norm, the problem is split-neighborly.

\begin{theorem}
\label{thm:box_ol}
The object localization problem where $S$ is an axis-symmetric box is $4$-split-neighborly. 
\end{theorem}

\begin{figure}
  \centering
  \includegraphics[width=\columnwidth]{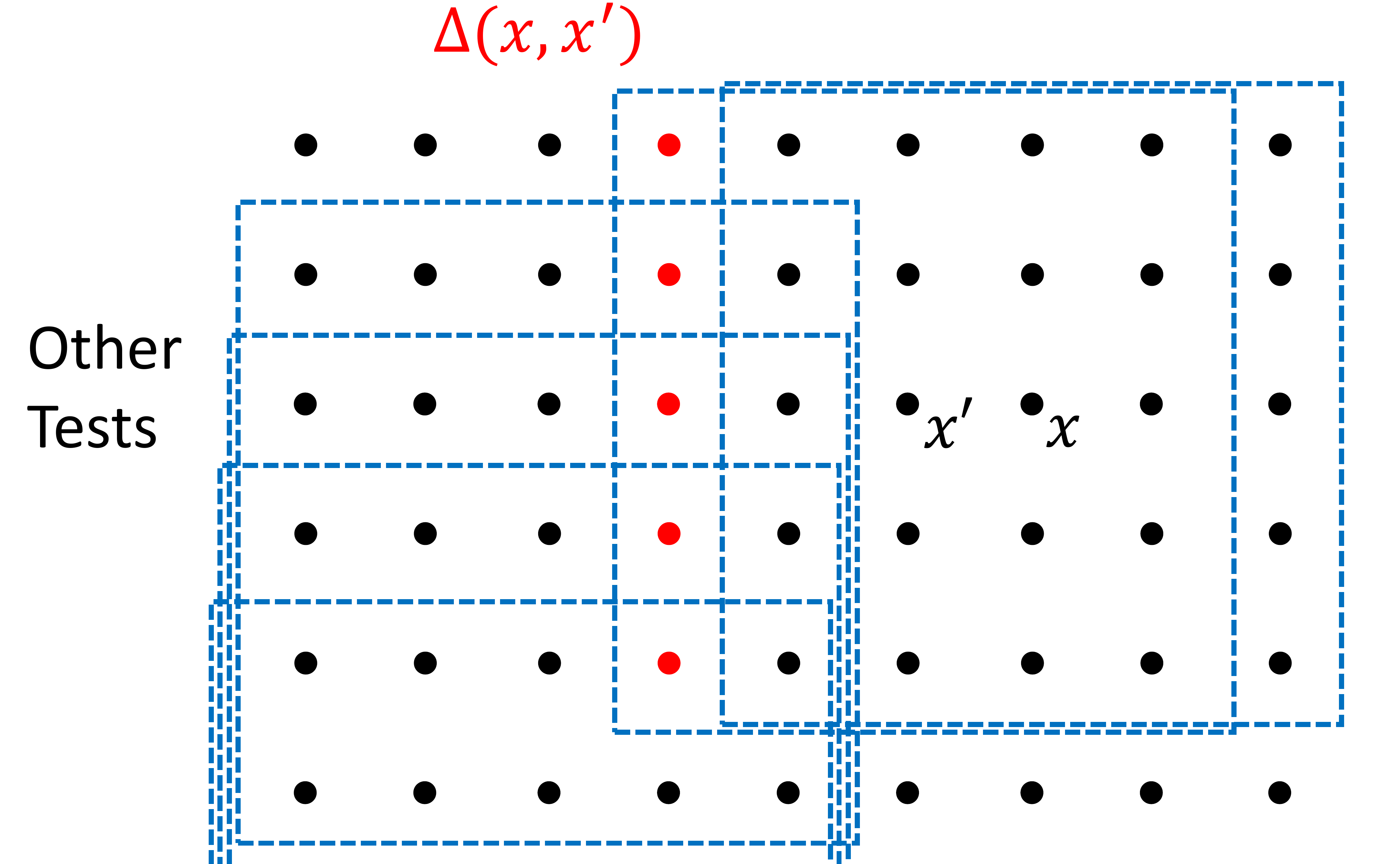}
\hfill
  \caption{For the object localization problem where $S$ is an axis-symmetric box, the \alphaSplitNeighborly\ graph has edges between adjacent points, for example $x$ and $x'$ in this figure. Thus, $\Delta(x,x')$ will be a flat box. This figure shows that for $d=2$, the problem is $3$-split-neighborly. In fact, the problem is $4$-split-neighborly for all $d$.}
  \label{box_ol}
\end{figure}

The proof is in the appendix. See Figure \ref{box_ol} for some intuition. Note that the value of $k$ for this problem is the largest cross-sectional volume of the box, which for $d\geq 2$, $k \geq \sqrt{n}$.

For axis-symmetric, axis-convex (weaker than convex) sets $S$, we have a dimension-dependent bound. By an axis-convex set $S$, we mean that if two points in $S$ differ in only one dimension, then all integral points between them are also in $S$.

\begin{theorem}
\label{thm:convex-shape}
If $S$ is a bounded, axis-symmetric, axis-convex set, the object localization problem is ${(4d+1)}$-split-neighborly. 
\end{theorem}

The proof is in the appendix and uses union bound with the partition of $\sX$ into $\sX^-$ and $\sX^+$. For this problem, $k$ must be at least the largest, axis-aligned ``shadow'' which for $d \geq 2$, $k \geq \sqrt{n}$.

\subsection{Discrete binary linear classifier}

Linear classifiers are a classic type of function where the output label is $y = \mathbf{1}[w \cdot x + b > 0]$. 
We consider the active learning setting where $\mathcal{H}$ is a set of $(w,b)$ pairs and $\mathcal{X}$ are points $x$ in the feature space. In fact, this is the problem covered previously by \citet{nowak2011geometry} where $w$ and $x$ take continuous values.

Here, we consider the setting of discrete linear classifiers, where $x \in \{0,1\}^d$ and $w \in \{-1, 0, 1\}^d$.

\begin{setting}[Discrete Binary Linear Classifier]
\label{ps:binary_linear}
$$\mathcal{H} \subseteq \{h_{b,w}\}_{b \in \mathbb{Z}, w \in \{-1, 0, 1\}^d}$$
$$\mathcal{X} = \{0,1\}^d$$
$$h_{b,w}(x) = \mathbf{1}[w \cdot x > b]$$
  with the following holding for all hypotheses ($w^{(+)}$ and $w^{(-)}$ are the number of positive and negative vector components for $w$, respectively):
$$w^{(+)} - b \leq r (w^{(-)} + b) - \frac{d}{8}$$
$$w^{(-)} + b \leq r (w^{(+)} - b - 1) - \frac{d}{8}$$
\end{setting}
Intuitively, $w^{(+)} - b$ is the maximum ``overshoot'' of the threshold and
$w^{(-)} + b$ is the maximum ``undershoot''. We require that the ratio between
these quantities is at most $r$ with the addition of an additive constant.

\begin{theorem}
\label{thm:binary_linear}
The discrete binary linear classifier problem is $\max(16,8r)$-split-neighborly.
\end{theorem}

The coherence parameter will be constant $c$ if there is some data distribution such that each hypothesis yields a balanced label distribution (at least probability $c$ of minority label). Intuitively, this is necessary since if the labels are very unbalanced, it may take many queries to even find a minority label. With this condition, GBS achieves $O(\log n)$ on the discrete binary linear classifiers problem.

On the other hand, the $k$-neighborly analysis does not work here, as before. In the special case where $d$ is divisible by $4$, $b=d/4 - 1$ and there are an equal number of $1$ and $0$ weights (just for making the calculation simpler), for $d \geq 4$, $k \geq \sqrt{n}$. See the Appendix for more details.
\section{Discussion and related work}
\label{sec:discussion}

The GBS algorithm, or more generally, choosing the test that maximizes the information gain, has several approximations and variants. The greedy information gain technique was introduced in \citet{mackay1992information} and used or extended in active learning \citep{jedynak2012twenty}, ranking learning \citep{chu2005extensions}, comparison based search \citep{karbasi2012comparison}, image segmentation \citep{maji2014active}, structured prediction \citep{sun2015active,luo2013latent}, group identification \citep{bellala2010extensions}, and graphical models \citep{zheng2012efficient}.
As the active learning survey of \citet{settles2012active} notes, ``all of the general query frameworks we have looked at contain a popular utility function that can be viewed as an approximation to [information gain] under certain conditions.'' Thus, greedy information gain is seen as the gold standard, and there has been significant work finding approximations and extensions. Our work  examines the other side of GBS and tries to understand that gap between GBS and the optimal solution.

A large body of literature exists on the analysis of GBS and close relatives in the noiseless and well-specified version of the sequential hypothesis testing problem, known as the optimal decision tree problem \citep{guillory2009average, kosaraju1999optimal,dasgupta2004analysis,chakaravarthy2007decision,adler2008approximating,chakaravarthy2009approximating,gupta2010approximation}. These analyses, which borrow ideas from submodular analysis, yield an \emph{average} cost \emph{ratio} of $O(\log n)$, where the average cost ratio for a method is defined as the ratio between the expected cost of the method and the expected optimal cost (note that this is significantly worse than an average query cost of $O(\log n)$). Furthermore, there exists a problem where GBS achieves a average cost ratio of $\Theta(\log n/\log\log n)$ (optimal is $\Theta(\log n)$ but GBS is $\Theta(\log^2 n / \log \log n)$), so the general upper bound for GBS is very close to tight \citep{dasgupta2004analysis}. In our work, we show that GBS achieves a \emph{constant factor} cost ratio, that is, within a constant factor of the optimal cost. In many natural settings such as linear classification, the hypothesis space is exponentially large in the dimension (i.e. $n = 2^{O(d)}$), so existing guarantees are $O(d)$ times the optimal, which itself is $O(d)$ for many problems. In our work, we prove in multiple settings that GBS achieves the asymptotically optimal query cost.

Other works extend the noiseless and well-specified assumptions to more general frameworks.
\citet{nowak2011geometry} provides a way to adapt GBS to the mis-specified case with only a constant factor increase in the query complexity that ensures GBS never performs worse than randomly querying tests (the naive approach). \citet{kaariainen2006active} provides a reduction from the noisy case to the noiseless case. There are two different noise settings which are handled separately in the literature, persistent noise (tests are not repeatable) and non-persistent noise (tests are repeatable). Earlier work \citep{nowak2009noisy, nowak2011geometry, naghshvar2012noisy} that has handled noise has addressed ~\emph{i.i.d.} noise with repeatable tests, where the outputs of the deterministic problem are flipped with a constant probability $p$.  In the case of non-persistent \emph{i.i.d.} noise, \citet{kaariainen2006active} presents a technique to reduce the noisy case to the deterministic case by repeatedly querying tests and using the majority vote, so that with high probability we attain the uncorrupted test result. Thus, while our work might appear to only handle the noiseless case, it actually handles the non-persistent noise case as well.


Theoretical explanations for the effectiveness of GBS are still incomplete.
Although GBS always achieves a cost ratio of $O(\log n)$ \citep{guillory2009average}, in the large hypothesis space regime, this factor could be very large.
Furthermore, there do exist problems for which GBS performs much worse than the optimal \citep{dasgupta2004analysis}.
These examples, however, tend to be contrived.  Anecdotally, from the sample problems in this paper, 
we found that GBS is effective for most ``natural'' problems.
In conclusion, we have made progress on characterizing this observation by introducing the $\nicefrac{1}{\alpha}$-split-neighborly condition,
which provably ensures that GBS achieves the asymptotically optimal query cost.

\subsubsection*{Acknowledgments}
\label{sec:acknowledgments}

This research was supported by NSF grant DGE-$1656518$.

\subsubsection*{References}
\let\bibsection\relax
\bibliography{bibliography}
\bibliographystyle{icml2017}

\clearpage
\clearpage

\section{Appendix}

\subsection{Split-neighborly proofs}

\begin{thmn}[\ref{split-neighborly}]
If a problem is $\nicefrac{1}{\alpha}$-split-neighborly and has a coherence parameter of $c$, for 

$$\beta = \min(c, \frac{1}{\nicefrac{1}{\alpha}+2})$$

GBS has a worst case query cost of at most $\frac{\log n}{-\log(1-\beta)}$ and GBS has an average query cost of at most $\frac{\log n}{H(\beta)}$ where $H(p)$ is the entropy of a $\text{Bernoulli}(p)$ random variable 

\end{thmn}
\begin{proof}
This theorem will follow from the next three lemmas.
\end{proof}

\begin{lem}[\ref{split-neighborly-lemma}]
If a problem is $\nicefrac{1}{\alpha}$-split-neighborly and has a coherence parameter of $c$, then for any $V \subseteq \sH$, $|V| \leq 1$ or there exists a test $x \in \sX$ such that 

$$\mathbb{E}_{h \in V}[h(x)] \in [\beta, 1-\beta]$$

where $\beta$ is defined as above.
\end{lem}
\begin{proof}
Fix a subset $V \subseteq \mathcal{H}$. Assume $|V| > 1$, otherwise we are done.

From the assumption, we have a coherence parameter of

$$
c \geq \beta
$$

From the definition, this means that there exists a probability distribution on the tests $P$ such that for any hypothesis $h$,

$$
\sum_{x \in X} P(x) h(x) \in [\beta, 1-\beta]
$$

Since this is true for all $h \in \mathcal{H}$, this is also true for all convex combinations. Thus,

$$
\mathbb{E}_{h \in V} [\sum_{x \in X} P(x) h(x)] \in [\beta, 1-\beta]
$$

$$
\sum_{x \in X} P(x) \mathbb{E}_{h \in V} [h(x)] \in [\beta, 1-\beta]
$$

For simplicity, define the split constant $S(x) = \mathbb{E}_{h \in V} [h(x)]$. Thus,

$$
\sum_{x \in X} P(x) S(x) \in [\beta, 1-\beta]
$$

There are two possibilities, either there exists a test $x$ such that 

$$
S(x) = \mathbb{E}_{h \in V} [ h(x) ] \in [\beta, 1-\beta]
$$

in which case, this is the exact conclusion statement and we are done, or that there exists no test with a split constant in $[\beta, 1-\beta]$. If there exists no test with a split constant in $[\beta, 1-\beta]$ but the weighted combination is in $[\beta, 1-\beta]$, then there exists tests $x$ and $x'$ such that $S(x) < \beta$ and $S(x') > 1-\beta$

Since the problem is $\nicefrac{1}{\alpha}$-split neighborly, there exists a graph over the tests that is strongly connected. Thus, there is a path from $x$ to $x'$. Since $S(x)<\beta$ and $S(x')>1-\beta$ and since $\forall x'' \in \sX: S(x'') \not\in [\beta,1-\beta]$, there must be an edge $(x_0, x_1)$ along the path where $S(x_0)< \beta$ and $S(x_1)> 1-\beta$. Thus,

$$
\Pr_{h \in V} [h(x_0)=1] = \mathbb{E}_{h \in V} [ h(x_0) ] < \beta
$$
$$
\Pr_{h \in V} [h(x_1)=1] = \mathbb{E}_{h \in V} [h(x_1)] > 1 - \beta
$$

Combining these two yields,

$$
\Pr_{h \in V} [h(x_0)=0 \wedge h(x_1)=1] > 1 - 2\beta
$$

Recall $\Delta(x_0,x_1) = \{h \in \mathcal{H}: h(x_0)=0, h(x_1)=1\}$

$$
\Pr_{h \in V} [h \in \Delta(x_0,x_1)] > 1 - 2\beta
$$
$$
\frac{|V \cap \Delta(x_0,x_1)|}{|V|}  > 1 - 2\beta
$$

Recall from the definition of $\beta$ that $\frac{1}{\nicefrac{1}{\alpha} + 2} \geq \beta$. Thus

$$1 - 2\beta \geq 1 - 2 \frac{1}{\nicefrac{1}{\alpha} + 2} = \frac{\nicefrac{1}{\alpha}}{\nicefrac{1}{\alpha} + 2} \geq \frac{\beta}{\alpha}$$

Thus,

$$
\frac{|V \cap \Delta(x_0,x_1)|}{|V|}   > \frac{\beta}{\alpha}
$$

For brevity, define $\Delta = \Delta(x_0,x_1)$. Since there is an edge $(x_0 , x_1)$ in the $\nicefrac{1}{\alpha}$-neighborly graph, for any subset including $V \cap \Delta \subseteq \Delta$, either $|V \cap \Delta| \leq 1$ or there exists a test $\hat{x}$ such that,

$$
\mathbb{E}_{h \in V \cap \Delta} [ h(\hat{x}) ] \in [\alpha, 1-\alpha]
$$

First, $|V \cap \Delta| \neq 0$, since $|V|>1$ and $\frac{|V \cap \Delta(x_0,x_1)|}{|V|}   > \frac{\beta}{\alpha}$. If $|V \cap \Delta| = 1$, then,
$\frac{|V \cap \Delta(x_0,x_1)|}{|V|} > \frac{\beta}{\alpha}$ and $|V|>1$ so $\frac{1}{2} \geq \frac{1}{|V|} > \frac{\beta}{\alpha} \geq \beta$. Since the hypotheses are identifiable, any pair of hypotheses yield a different result on some test, so we can always find a test with a split constant of at least $\frac{1}{|V|}$, and this implies the result of the theorem.

In the other case, where $|V \cap \Delta| > 1$, we have all the necessary pieces and it's just a matter of crunching the algebra.

$$\mathbb{E}_{h \in V}[h(\hat{x})] = \frac{\sum_{h \in V} h(\hat{x})}{|V|}$$
$$\geq \frac{\sum_{h \in V \cap \Delta} h(\hat{x})}{|V|}$$
$$\geq  \frac{\beta}{\alpha} \frac{\sum_{h \in V \cap \Delta} h(\hat{x})}{|V \cap \Delta|}$$
$$\geq \frac{\beta}{\alpha} \mathbb{E}_{h \in V \cap \Delta}[h(\hat{x})]$$
$$\geq \frac{\beta}{\alpha} \alpha = \beta$$

Additionally,

$$\mathbb{E}_{h \in V}[h(\hat{x})] = \frac{\sum_{h \in V} h(\hat{x})}{|V|}$$
$$= \frac{\sum_{h \in V \cap \Delta} h(\hat{x}) + \sum_{h \in V \setminus \Delta} h(\hat{x})}{|V|}$$
$$\leq  \frac{(1-\alpha) |V \cap \Delta| + \sum_{h \in V \setminus \Delta} h(\hat{x})}{|V|}$$
$$\leq  \frac{(1-\alpha)|V \cap \Delta| + |V| - |V \cap \Delta|}{|V|}$$
$$\leq 1 - \alpha \frac{|V \cap \Delta|}{|V|}$$
$$\leq 1 - \alpha \frac{\beta}{\alpha} = 1 - \beta$$

Thus, we have that

$$\mathbb{E}_{h \in V}[h(\hat{x})] \in [\beta, 1 - \beta]$$

which is the conclusion of the lemma.

\end{proof}

\begin{lemma}
If, for any $V \subseteq \sH$, $|V| \leq 1$ or there exists a test $x \in \sX$ such that 

$$\mathbb{E}_{h \in V}[h(x)] \in [\beta, 1-\beta]$$

then GBS has a worst case query cost of at most $\frac{\log n}{\log(\frac{1}{1-\beta})}$
\end{lemma}
\begin{proof}
After $m$ queries, there are at most $\max(1,(1-\beta)^m n)$ remaining hypotheses since greedy will choose a test with a split constant of at least $\beta$ (a split with respect to the hypotheses without a prior) and will terminate when there is a single hypothesis. Thus, when $(1-\beta)^m n \leq 1$, the algorithm must have terminated. Rearranging, we see that when $m \geq \frac{\log n}{\log(\frac{1}{1-\beta})}$ the algorithm must have terminated. This means that the worst case query cost must be at most $\frac{\log n}{\log(\frac{1}{1-\beta})}$.
\end{proof}

\begin{lemma}
If, for any $V \subseteq \sH$, $|V| \leq 1$ or there exists a test $x \in \sX$ such that 

$$\mathbb{E}_{h \in V}[h(x)] \in [\beta, 1-\beta]$$

 then GBS has an average query cost of at most $\frac{\log n}{H(\beta)}$ where $H(p)$ is the entropy of a $\text{Bernoulli}(p)$ random variable 
\end{lemma}
\begin{proof}

Define $H(p)$ as the entropy of a Bernoulli random variable with probability $p$. 

\begin{equation}
f(V) = \mathbb{E}[\text{average queries remaining while at subset }V]
\end{equation}

We will prove by induction on increasing subsets that 

\begin{equation}
f(V) \leq \frac{\log(|V|)}{H(\beta)}
\end{equation}

Note that the base case is that $f(\{h\}) = 0$ because we are done when there is just one hypothesis left. Note that this suffices to show that the total runtime is $\log(n)/ H(\beta)$ because $|V|=n$ at the beginning of the algorithm.




Let $A$, $B$ be a partition of $V$ based on a test split. Without loss of generality, let $|A| \leq |B|$, so $|A| \leq \nicefrac{1}{2} |V|$. Based on the recursive definition of cost and there is a test with a split constant of at least $\beta$ (so GBS will choose a test with a split constant of at least $\beta$),

$$f(V) \leq \max_{A,B, \nicefrac{|A|}{|V|} \in [\beta,\nicefrac{1}{2}] } \frac{|A|}{|V|} f(A) + \frac{|B|}{|V|} f(B) + 1$$

From the induction hypothesis,

$$\leq \max_{...} \frac{|A|}{|V|} \frac{\log|A|}{H(\beta)} + \frac{|B|}{|V|} \frac{\log|B|}{H(\beta)} + 1$$

$$\leq \frac{ \max_{...} \frac{|A|}{|V|} \log|A| + \frac{|B|}{|V|} \log|B| + H(\beta) }{H(\beta)}$$

$$\leq \frac{\max_{... } \frac{|A|}{|V|} \log\frac{|A|}{|V|} + \frac{|B|}{|V|} \log\frac{|B|}{|V|} + H(\beta) + \log|V| }{H(\beta)}$$

$$\leq \frac{\max_{... } -H(\frac{|A|}{|V|}) + H(\beta) + \log|V|  }{H(\beta)}$$

Note that since $\nicefrac{|A|}{|V|} \in [\beta,\nicefrac{1}{2}] $ (the condition of the max), $H(\frac{|A|}{|V|})\geq H(\beta)$. Thus, the $\max$ is non-positive, and thus,

$$f(V) \leq  \frac{\log(|V|)}{H(\beta)} $$

Thus, we have proved the statement by induction and this suffices to show that the total runtime is at most $\log(n) / H(\beta)$.

\end{proof}

\begin{prop}[\ref{k-neighborly-to-split-neighborly}]
If a problem is $k$-neighborly and has a uniform prior, then the problem is $k$-split-neighborly.
\end{prop}
\begin{proof}
In the case that $k=1$, $|\Delta(x,x')|=1$ so $|V|\leq 1$ so the problem is $1$-split-neighborly. Assume $k>1$. Note that any set of hypotheses must have a test that distinguishes at least one of the hypotheses (otherwise the hypotheses are the same). If two points $x$ and $x'$ in the $k$-neighborly graph have an edge between them, then $|\Delta(x,x') \cup \Delta(x',x)| \leq k$, which implies $|\Delta(x,x')| \leq k$, and thus either $|\Delta(x,x')| \leq 1$ or there is a test with a $1/k$ split constant and thus there is an edge from $x$ to $x'$ in the $k$-split-neighborly graph. By a similar argument, there is also an edge from $x'$ to $x$. Since the $k$-neighborly graph is connected and each edge corresponds to a bidirectional edge in the $k$-split-neighborly graph, the $k$-split-neighborly graph is strongly connected and thus the problem is $k$-split-neighborly.
\end{proof}


\subsection{Value of $k$}
\label{sec:k_big}

\subsubsection{Disjunctions}
For the disjunctions problem, for $m \geq 2, d \geq 2m$,

$$n = \sum_{i=1}^m \binom{d}{i}$$
$$k \geq \sum_{i=1}^m \binom{d-1}{i-1}$$

$$k \geq 1 + \sum_{i=1}^{m-1} \binom{d-1}{i}$$

$$k^2 - n \geq 1 + 2 \sum_{i=1}^{m-1} \binom{d-1}{i} + (\sum_{i=1}^{m-1} \binom{d-1}{i})^2$$ $$ - \sum_{i=1}^{m-1} \binom{d}{i} - \binom{d}{m}$$

Note that $2 \binom{d-1}{i} \geq \binom{d}{i}$ since $i \leq m-1 \leq d/2$.

$$k^2 - n \geq 1 + (\sum_{i=1}^{m-1} \binom{d-1}{i})^2 - \binom{d}{m}$$
$$\geq \binom{d-1}{m-1}^2 - \binom{d}{m}$$
$$\geq \binom{d-1}{m-1} ( \binom{d-1}{m-1} - d/m)$$
Since $m \geq 2$,
$$\geq \binom{d-1}{1} - d/2$$
$$\geq d/2 - 1$$
$$\geq m - 1$$
$$\geq 0$$

Thus, $k^2 - n \geq 0$ and so $k \geq \sqrt{n}$.

\subsubsection{Monotonic CNF}

Note that $n = |\sH| = \frac{1}{l!} \binom{d}{m,m,...,m,d-lm}$. All of the bit strings with strictly less than $l$ ones will be trivially connected in the $k$-neighborly graph, because they yield $0$ on all hypotheses. However, the closest test to connect them to the rest of the graph is the bit string $1^l0^{d-l} \in \mathcal{X}$, which disagrees on $\binom{d-l}{m-1,m-1,...,m-1,d-lm} \leq k$ hypotheses. We examine the case where $d \geq 2ml$ and $m \geq 2$.

For the monotonic CNF formulas, recall that 

$$n = |\sH| = \frac{1}{l!} \binom{d}{m,m,...,m,d-lm}$$
$$k \geq \binom{d-l}{m-1,m-1,...,m-1,d-lm}$$

For $d \geq 2ml$ and $m \geq 2$, $k \geq \sqrt{n}$.

$$\binom{d-l}{m-1,m-1,...,m-1,d-lm} \leq k$$

and 

$$n = \frac{1}{l!} \binom{d}{m,m,...,m,d-lm}$$
$$= \frac{1}{l!} \frac{d!}{(m!)^l (d-lm)!}$$
$$= \frac{(d-l)!}{(m-1)!^l (d-lm)!} \frac{1}{m^l} \frac{d! (d-2l)!}{(d-l)!^2} \frac{(d-l)!}{l! (d-2l)!}$$
$$\leq k \frac{1}{m^l} \frac{d! (d-2l)!}{(d-l)!^2} \binom{d-l}{l}$$

Since $d \geq 2ml \geq 4l$,

$$n \leq k \frac{2^l}{m^l} \binom{d-l}{l}$$

Since $d-l \geq 2l(m-1)$ and $m \geq 2$

$$n \leq k \binom{d-l}{l(m-1)}$$
$$n \leq k \binom{d-l}{m-1,m-1,...,m-1,d-lm}$$
$$n \leq k^2$$
$$k \geq \sqrt{n}$$

\subsubsection{Discrete Linear Classifier}

Recall that we are in the special case where $d$ is divisible by $4$, $b=d/4 - 1$ and there are an equal number of $1$ and $0$ weights ($d/2$).

All tests with fewer than $d/4$ 1's will yield a result of $0$ for all hypotheses. The test with the next fewest hypotheses that yield $1$ will be a test with exactly $d/4$ 1's. Thus, $k$ is at least the number of such hypotheses that yield $1$.

$$n = \binom{d}{d/2}$$
$$k \geq \binom{3d/4}{d/4}$$

For simplicity, define $c=d/4$.

$$\frac{n}{k^2} \leq \frac{ \binom{4c}{2c} }{  \binom{3c}{c}^2  }$$

$$ = \frac{ (4c)! c! c!}{ (3c)! (3c)! }$$

Note that we have the common Stirling's approximation,

$$\sqrt{2 \pi} n^{n+1/2} e^{-n} \leq n! \leq e n^{n+1/2} e^{-n}$$

Thus,

$$\frac{n}{k^2} \leq \frac{e^3 (4c)^{4c+1/2} c^{c+1/2} c^{c+1/2} e^{-6c}}{2\pi (3c)^{3c+1/2} (3c)^{3c+1/2} e^{-6c}}$$

$$ = \frac{2 e^3 \sqrt{c} (4c)^{4c} c^c c^c}{6 \pi (3c)^{3c} (3c)^{3c}}$$
$$ = \frac{e^3 \sqrt{c} 4^{4c}}{3 \pi 3^{3c} 3^{3c}}$$
$$ = \frac{e^3}{3 \pi} \sqrt{c} (\frac{256}{729})^c$$
$$ \leq 1$$

for $c \geq 1$.

Thus, for $d \geq 4$,

$$\frac{n}{k^2} \leq 1$$
$$k \geq \sqrt{n}$$


\subsection{Necessity of Dependencies}

\subsubsection{Linear classifiers on convex polygon data pool}

\begin{figure}[t!]
\centering
  \includegraphics[width=.9\linewidth]{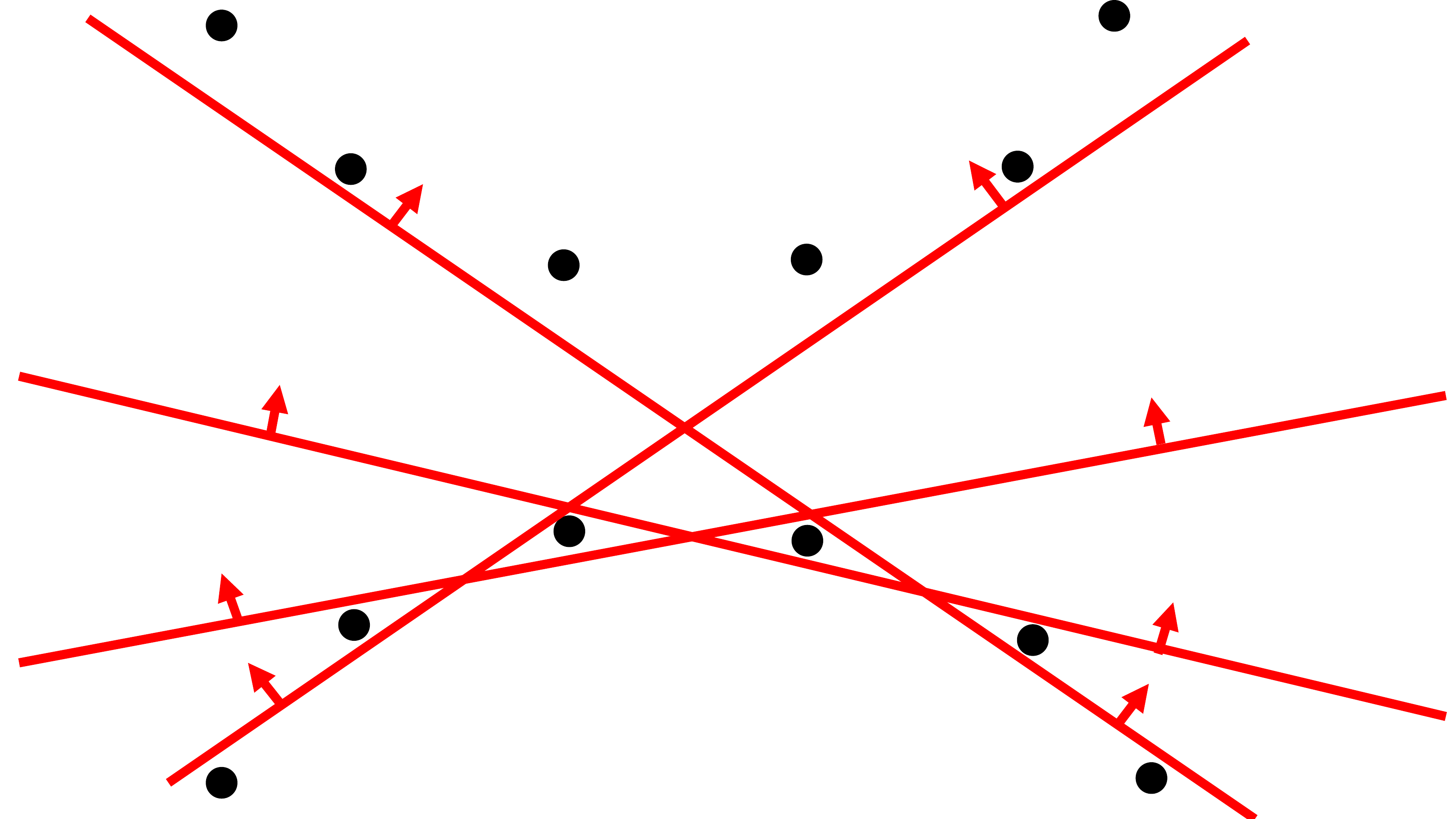}
  \caption{A counterexample that shows a non-convex data pool need not be split-neighborly. Note that we can at most split off $1$ of the $n$ hypotheses by querying one of the points from the lower half. However, the problem has coherence close to $1/2$ and thus it cannot be \alphaSplitNeighborly\ for constant $\alpha$.}
 \label{fig:lc-no-sn}
\end{figure}

For arbitrary data points where the points are not the vertices of a convex polygon, the linear classifier problem is not \alphaSplitNeighborly\ for constant $\alpha$. A counter-example is shown in Figure \ref{fig:lc-no-sn}. 

\subsubsection{Disjunctions}
The linear dependence on $m$ for the disjunctions is necessary because of the case where $d=m+1$, and $|\sH|=d$ (each $h \in \sH$ lacking one variable). In this case, there are no tests with split constants of $\frac{1}{m}$, so the problem cannot be better than $(m-2)$-split-neighborly (recall coherence $c=1/2$).

\subsubsection{Monotonic CNF}

For the monotonic CNF problem, the linear dependence on $m$ is necessary because of the case where $l=1$, $d=m+1$, and $|\sH|=d$ (each $h \in \sH$ lacking one variable). In this case, there are no tests with split constants of $\frac{1}{m}$, so the problem cannot be better than $(m-2)$-split-neighborly (recall coherence $c=1/2$). Furthermore, the linear dependence on $l$ is necessary because of the problem where $m=1$, $d=l+1$, and $|\sH|=d$ (each $h \in \sH$ lacking one variable). For this problem, there are no tests with split constants of $\frac{1}{l}$, so the problem cannot be better than $(l-2)$-split-neighborly. Thus, although the linear dependence on $m$ and $l$ is necessary, it may be possible to improve the constants.

\subsubsection{Object Localization}

For object localization with the axis-symmetric, axis-convex set $S$, the dependence on $d$ is necessary because if we use the set $S = \{j e_i: |j| \leq l, 1 \leq i \leq d\}$ and consider the set of hypotheses, $\{\pm l e_i : 1 \leq i \leq d\}$, the problem has no test with split constant of $\frac{1}{2d-1}$ but has coherence $c=1/2$, so it can't be $(2d-3)$-split-neighborly.



\subsection{Monotonic CNF}

\begin{thmn}[\ref{thm:cnf}]
The Conjunction of Disjunctions problem is $(m+1+3(l-1))$-split-neighborly.
\end{thmn}
\begin{proof}

We prove this theorem by induction on $l$. First, for the base case $l=1$.

The test graph has an edge from $x$ to $x'$ if $||x - x'||_1 = 1$ (the bit strings differ in one location).

Let $x^+$ be the value of $x$ or $x'$ with more $1$'s (and let $x^-$ be the other one). Note that $|\Delta(x^+,x^-)| = 0$ so there is a directed edge $(x^+,x^-)$.

For the other direction, fix a subset $V \subseteq \Delta(x^-,x^+)$. Without loss of generality, let $x^+$ and $x^-$ differ in the first coordinate so $x^+_1 = 1$ and $x^-_1=0$ and $\forall i>1: x^+_i = x^-_i$. 

  For a proof by contradiction, the problem is not $(m+1)$-split-neighborly so that $|V|>1$ and there is no test $x$ such that $\mathbb{E}_{h \in V}[h(x)] \in [q, 1 - q]$, where $q=1/(m+1)$. 

Let 

$$\sX^+ = \{x \in \sX: \Pr_{h \in V}[h(x)=1] > 1-q\},$$
$$\sX^- = \{x \in \sX: \Pr_{h \in V}[h(x)=1] < q\} = \sX - \sX^+.$$

Let $x'$ be the the element of $\sX^-$ with the fewest $0$'s and let the $0$'s be at indices $Z$ (note $1 \in Z$). If $|Z|<m$, then $h(x')=1$ for all $h$ since the disjunctions have $m$ variables. But since $x' \in \sX^-$, which is a contradiction.

Define $\{x^{(j)}\}_{j \in Z}$ to be the test resulting changing the $j^{th}$ bit of $x'$ to a $1$. By the minimal definition of $x'$, $\forall j \in Z: x^{(j)} \in \sX^+$. 

Suppose $|Z| > m$. Take a subset $Z' \subseteq Z$ such that $|Z'|=m+1$. Then, from the definition of $\sX^+$ and $\sX^-$, $\Pr_{h \in V}[h(x')=0 \wedge \forall j \in Z': h(x^{(j)})=1] > 1 - (m+1)q \geq 0$, which means $\Pr_{h \in V}[h \text{ includes variables } Z'] > 0$. Therefore, there is a disjunction with at least $m+1$ variables, which is a contradiction.

Thus, $|Z| = m$, so there is only one hypothesis such that $h(x')=0$, the hypothesis with variables at $Z$. So $1/|V| > 1-q$ (by definition of $\sX^-$), which implies $|V|=1$ since $q\leq 1/2$, which is a contradiction. Thus, by contradiction, the problem with $l=1$ is $(m+1)$-split-neighborly.
For $l>1$, we proceed by induction. We can define the graph as above, define $\sX^-$ and $\sX^+$ as above, and $x'$ and $Z$ as above. The same argument goes through that $|Z| = m$. Thus, $(1-q)$ proportion of the hypotheses have a disjunction with variables at the indices $Z$. These hypotheses are simply another copy of the problem with $l-1$ conjunctions and $d-m$ variables. Since that problem has $1/2$ coherence and is $m+1+3(l-2)$-splittable (by induction hypothesis), there exists some test with a split constant of $\frac{1}{m+1+3(l-2)+2}$ for a total split constant on the original problem of 

$$(1-q)\frac{1}{m+1+3(l-2)+2} = \frac{1}{m+1+3(l-1)}$$

Thus, the problem is $m+1+3(l-1)$-split-neighborly by induction.
\end{proof}

\subsection{Box Object Localization}

\begin{thmn}[\ref{thm:box_ol}]
The object localization problem where $S$ is a box is $4$-split-neighborly. 
\end{thmn}

Notationally, refer to $z_h$ as the integer vector for the hypothesis $h$ and $z_{h,i}$ to be its $i^{th}$ component.

We begin by fixing two tests $x$ and $x'$ such that $||x-x'||_1 = 1$. Without loss of generality, let $x'-x = e_1$ where $e_1$ is the $1^{st}$ elementary vector. Since the box is axis symmetric, there exists radii $r_i \geq 0$ such that $x - z_h \in S \leftrightarrow \forall i: |x_i - z_{h,i}| \leq r_i$. Without loss of generality, assume $x = (r_1, 0, 0, ..., 0)$ and $x' = (r_1+1, 0, 0, ..., 0)$. Recall $\Delta(x,x') = \{h: h(x)=0 \wedge h(x')=1\}$, this implies that $\Delta(x,x') = \{h: z_{h,1} = 0 \wedge \forall i>1: |z_{h,i}| \leq r_i\}$. We will begin by fixing a subset $V \subseteq \Delta(x,x')$. As in all the application proofs, we will start by assuming by contradiction that there is no test with a split constant in the range $[q,1-q]$ where $q=1/4$. We will use this contradiction to show that the size of $V$ is small, so that there is in fact a test with a split constant $q$ which is a contradiction.

\subsubsection{Majority Element}

Fix a dimension $i$. Examine the tests $X_i = \{j e_i : j=0,..,2r_i+1\}$ and note that for $h \in V \subseteq \Delta(x,x')$, $h(j e_i) = \mathbf{1}[z_{h,i} \geq j - r_i]$.

By the contradiction assumption,
$$\mathbb{E}_{h \in V}[h(j e_i)] \not\in [q, 1-q]$$

$$\Pr_{h \in V}[z_{h,i} \geq j - r_i] \not\in [q,1-q]$$

Since $\Pr_{h \in V}[z_{h,i} \geq -r_i ] = 1$ and $\Pr_{h \in V}[z_{h,i} \geq r_i+1] = 0$, there must be some integer $m_i$ such that

$$\Pr_{h \in V}[z_{h,i} \geq m_i ] > 1-q$$
$$\Pr_{h \in V}[z_{h,i} \geq m_i+1 ] < q$$

which implies that

$$\Pr_{h \in V}[z_{h,i} = m_i ] > 1-2q$$

Define thus, there exists a vector $m$ such that there is a $1-2q$ probability that an hypothesis' $i^{th}$ component matches $m$.

\subsubsection{Side Splits}

Intuitively, we will create a sequence of tests that each remove at least half of the elements with the $i^{th}$ component not equal to $m$. For each test in the sequence, the probability that the test yields $1$ over the hypotheses in $V$ must be greater that $1-q$ so we can prove that there aren't many elements that disagree with $m$ at any component.

Here we recursively define sets $S_i$, $B_i$, and $A_i$. $S_i$ will be defined in terms of $B_{i}$ and $B_i$ will be defined in terms of $S_{i-1}$.

Define $S_0 = V$ and for $i>1$, $S_i = S_{i-1} - B_{i}$. Noting that we could reflect the $i^{th}$ component about $m_i$, without loss of generality, suppose that 

$$\Pr_{h \in S_i}[z_{h,i} > m_i] \geq \Pr_{h \in S_i}[z_{h,i} < m_i]$$

Define $B_i = \{h \in S_{i-1}: z_{h,i} > m_i\}$ and $A_i = \{h \in S_{i-1}: z_{h,i} < m_i\}$

Note that $|B_i| \geq |A_i|$.

Further, there is a test $x^{(i)} = (-r_1,...,-r_i,0,...0)$ such that $h(x^{(i)}) = 1 \leftrightarrow h \in S_i$ and thus by the contradiction assumption,

$$\frac{|S_i|}{|S|} \not\in [q,1-q]$$

However, since $\Pr_{h \in V}[z_{h,i} = m_i ] > 1-2q$, $|B_i|/|V| < 2q$. We now prove by induction that $|S_i|/|V| > 1-q$. The base case is that $|S_1|/|V|=1 > 1-q$. As long as $q \leq 1/4$, since $|S_{i-1}|/|V| > 1-q$ and $|B_i|/|V| < 2q$, $|S_i|/|S| > 1-3q \geq q$ (since $q=1/4$)and thus by the contradiction assumption $|S_i|/|S| > 1-q$.

Note that the $B_i$ are disjoint because

$$B_i \subseteq S_i = V - B_1 - B_2 - ... - B_{i-1}$$

$$|S_d| > (1-q) |V|$$
$$|V - \bigsqcup_{i=1}^d B_i| > (1-q) |V| $$
$$|V| - \sum_{i=1}^d |B_i| > (1-q) |V|$$ 
$$q |V|  > \sum_{i=1}^d |B_i|$$ 

Define the set of elements $M' \subseteq V$ as the points with a component not equal to $m$. This is the union of all $A_i$ and $B_i$,

$$ |M'| = |\bigcup_{i=1}^d A_i \cup \bigcup_{i=1}^d B_i|$$
$$ \leq \sum_{i=1}^d |A_i| + \sum_{i=1}^d |B_i|$$
$$ \leq 2 \sum_{i=1}^d |B_i|$$
$$ < 2q |V| $$

Also note that $|M'| \geq |V| - 1$ since there can only be one element that doesn't disagree with any element of $m$. Thus,

$$|V| - 1 < 2q |V|$$
$$|V| < \frac{1}{1-2q}$$

Since $q \leq 1/3$, then this implies $|V| < 3$ so there is a test with a split of $1/3$, which is a contradiction. So in a proof by contradiction, the problem is $4$-split-neighborly.

\subsection{Convex, axis-symmetric Shape Object Localization}

\begin{thmn}[\ref{thm:convex-shape}]
If $S$ is a bounded, axis-symmetric, axis-convex shape, the object localization problem is $(4d+1)$-split-neighborly. 
\end{thmn}
\begin{proof}
Let the test graph has an edge from $x$ to $x'$ if $||x - x'||_1 = 1$. 

Fix a subset $V \subseteq \Delta(x,x')$. Without loss of generality, let $x'=0^d$. $V \subseteq \Delta(x,x') \subseteq \{h: h(x')=1\} = \{h: z_h - x' \in S\} = \{h: z_h \in S\}$

For a proof by contradiction, the problem is not $4d+1$-split-neighborly so that $|V|>1$ and there is no test $x$ such that $\mathbb{E}_{h \in V}[h(x)] \in [q, 1 - q]$, where $q=1/(4d+1)$. 

Let 

$$\sX^+ = \{x \in \sX: \Pr_{h \in V}[h(x)=1] > 1-q\}$$
$$\sX^- = \{x \in \sX: \Pr_{h \in V}[h(x)=1] < q\} = \sX - \sX^+$$

Note that $x' = 0^d \in \sX^+$ since $V \subseteq \{h: h(x')=1\}$.

Fix a dimension $i$. Examine the set of tests $\{je_i: j \in \mathbb{Z}\}$. From above, $0e_i \in \sX^+$. Further, since $V \subseteq \{h: z_h \in S\}$ and since $S$ is bounded, there exists some $B \in \mathbb{Z}$ such that $\pm Be_i \in \sX^-$. Thus there exists some $c_1 \leq 0, c_2 \geq 0$ such that $(c_1 - 1)e_i \in \sX^-$, $c_1e_i \in \sX^+$, $c_2e_i \in \sX^+$, $(c_2 + 1)e_i \in \sX^-$. From the definition of $\sX^+$ and $\sX^-$,

$$\Pr_{h \in V}[h((c_1 - 1)e_i)=0, h(c_1e_i)=1,...$$
$$h(c_2e_i)=1, h((c_2 + 1)e_i)=0] > 1-4q$$

Define $S_{l} = \{s_{-i} : s_i=l, s \in S\}$ to be the slices of $S$ along axis $i$ at location $l$. Therefore, $h(je_i)=1 \leftrightarrow z_{h,-i} \in S_{z_{h,i}-j}$.

Note that $S_{-l} = S_{l}$ since the shape $S$ is axis symmetric. Combining these three facts,

$$\Pr_{h \in V}[z_{h,-i} \in S_{|z_{h,i}-c_1|} \cap S_{|z_{h,i}-c_2|} \setminus ...$$ $$ \setminus(S_{|z_{h,i}-(c_1-1)|} \cup S_{|z_{h,i}-(c_2+1)|} )]> 1-4q$$

Note that for $l' > l \geq 0$, $S_{i,l'} \subseteq S_{i,l}$ because of axis-convexity. To see this, suppose there was $t \in S_{i,l'} \setminus S_{i,l}$, then there would be three elements $s^{(-1)},s^{(0)},s^{(1)}$ such that $s^{(j)}_{-i} = t$ and $s^{(-1)}_i=-l',s^{(0)}_i=l,s^{(1)}_i=l'$, which would imply $s^{(-1)} \in S,s^{(0)} \not\in S,s^{(1)} \in S$ which contradicts axis-convexity.

Thus, in order for the set composed of slices of $S$ in the equation above to be non-empty,

$$|z_{h,i}-c_1|, |z_{h,i}-c_2| < |z_{h,i}-c_1+1|, |z_{h,i}-c_2-1|$$

it must be the case that $z_{h,i} = \frac{c_1 + c_2}{2} \in \mathbb{Z}$ which we define to be $m_i$. So,

$$\Pr_{h \in V}[z_{h,i} = m_i] > 1-4q$$

Repeating this argument for all dimensions and combining,

$$\Pr_{h \in V}[ \forall i: z_{h,i} = m_i] > 1 - 4dq$$

There is only one such element $z_h = m$ so

$$\frac{1}{|V|} > 1 - 4dq = 1 - 4d \frac{1}{4d+1} = \frac{1}{4d+1}$$

So $|V| < 4d+1$ so there must be a split of at least $q$ which is a contradiction.

\end{proof}
\subsection{Discrete Binary Linear Classifiers}

\begin{thmn}[\ref{thm:binary_linear}]
The discrete binary linear classifier problem is $\max(16,8r)$-split-neighborly.
\end{thmn}

Define $q = \min(\frac{1}{16},\frac{1}{8r})$

Recall that for the Discrete Binary Linear Classifier case, we have hypotheses as a pair of vectors and threshold $h = (w_h, b_h) \in \{-1, 0, 1\}^d \times \mathbb{Z}$ and tests as vectors $\{0,1\}^d$. Recall $h(x) = \mathbf{1}[w_h \cdot x > b_h]$.

From the problem setting of Discrete Binary Linear Classifiers, we know that,

$$w_h^{(+)} - b \leq r (w_h^{(-)} + b) - \frac{d}{8}$$
$$w_h^{(-)} + b \leq r (w_h^{(+)} - b - 1) - \frac{d}{8}$$

Recall $w^{(+)}$ is the number of positive elements of $w$ and $w^{(-)}$ is the number of negative elements. Notationally $w_{h,i}$ refers to the $i^{th}$ component of $w_h$.

\subsubsection{Key Lemma and its Sufficiency}

We will first state a lemma and then prove that it implies the problem stated stated.

\begin{lemma}
\label{binary_linear_key_lemma}
Define
$$
x^{(0)} = (0, 0, ..., 0)
$$
$$
x^{(1)} = (1, 0, ..., 0)
$$
$$
H' = \{h \in H: h(x^{(0)} = 0 \wedge h(x^{(1)}) = 1 \wedge ... $$ $$\wedge w_h^{(+)} \leq r w_h^{(-)} - \frac{d}{8} \wedge ...$$
$$...\wedge w_h^{(-)} \leq r (w_h^{(+)} - 1) - \frac{d}{8}\}
$$

For any subset $V \subset H'$, there exists a test $x$ such that $\mathbb{E}_{h \in V}[h(x)] \in [q, 1-q]$
\end{lemma}

\subsubsection{Proof of Theorem \ref{thm:binary_linear} from Lemma \ref{binary_linear_key_lemma}}
We will prove Theorem \ref{thm:binary_linear} by a reduction to Lemma \ref{binary_linear_key_lemma}. To show that the problem is $\nicefrac{1}{\alpha}$-split-neighborly, we need to show that for two tests with $x$ and $x'$ with $||x-x'||_1 = 1$ that for any subset $V \subseteq \Delta(x,x') = \{h: h(x)=0 \wedge h(x')=1\}$, that $|V| \leq 1$ or there exists a test $\hat{x}$ such that 

$$\Pr_{h \in V}[h(\hat{x}) = 1] \in [q,1-q]$$

Note that by permuting the indices of $x$ and $x'$, we can make the first index the one that is different between $x$ and $x'$. Additionally, for the remaining indices we can flip the $0$'s and $1$'s of the test so long as we flip the non-zero entries of $w_h$ at that same position, and change $b_h$ accordingly. We flip the bits so that $x$ becomes $x^{(0)}$ and $x'$ becomes $x^{(1)}$.

Note that $h(x^{(0)}) = 0$ implies that $0 \leq b_h$. Further note that, $h(x^{(1)}) = 1$ implies that $w_{h,1} > b_h$. Thus, the only possibility is that $w_{h,1} = 1$ and $b_h = 0$. 

Let $T_{+-}$ denote the number of flips from positive to negative weights and let $T_{-+}$ denote the number of flips from negative to positive. Then, the weights for the new (reduction) problem will be

$$w_{new}^{(+)} = w^{(+)} + T_{-+} - T_{+-}$$
$$w_{new}^{(-)} = w^{(-)} + T_{+-} - T_{-+}$$
$$0 = b_{new} = b - T_{+-} + T_{-+}$$

From the last equation, $b = T_{+-} - T_{-+}$. Thus,

$$w_{new}^{(+)} = w^{(+)} - b$$
$$w_{new}^{(-)} = w^{(-)} + b$$

Since, 

$$w^{(+)} - b \leq r (w^{(-)} + b) - \frac{d}{8}$$
$$w^{(-)} + b \leq r (w^{(+)} - b - 1) - \frac{d}{8}$$

then,

$$w_{new}^{(+)} \leq r w_{new}^{(-)} - \frac{1}{8}d$$
$$w_{new}^{(-)} \leq r (w_{new}^{(+)} - 1) - \frac{1}{8}d$$

We can see that the hypothesis conditions for the original theorem imply that $\Delta(x^{(0)},x^{(1)})$ is a subset of the hypotheses that satisfy the conditions based on $w_{new}^{(-)}$ and $w_{new}^{(+)}$ so Lemma \ref{binary_linear_key_lemma} implies the binary linear classifier is $\nicefrac{1}{q}$-split-neighborly which means $\max(16,8r)$-split-neighborly.

\subsubsection{Proof of Lemma \ref{binary_linear_key_lemma}}

The remainder of this is devoted to proving Lemma \ref{binary_linear_key_lemma}

We begin by fixing a subset $V \subseteq H'$. As in all the application proofs, we will start by assuming by contradiction that there is no test with a split constant in the range $[q,1-q]$. We will use this contradiction to show that the size of $V$ is small.

Recall that $b_h=0$ for all hypotheses in the reduced problem and $w_{h,1}=1$. This follows from the fact that $h(x^{(0)}) = 0$ and $h(x^{(1)}) = 1$.

\subsubsection{Majority Vector}

Let $e_i$ be an elementary vector with all entries $0$ except for the $i^{th}$ entry which is $1$.

\begin{lemma}
There exists a vector $m \in \{-1, 0, 1\}^d$ such that $\forall i: m_i=0: \Pr_{h \in V}[w_{h,i} = m_i] \geq 1-2q$ and $\forall i: m_i\neq0: \Pr_{h \in V}[w_{h,i} = m_i] \geq 1-q$

\end{lemma}
\begin{proof}

By the contradiction assumption, there isn't a test with a split constant greater than $q$,

$$\Pr_{h \in V}[ w_h \cdot e_i > b_h ] \not\in [q, 1-q]$$
$$\Pr_{h \in V}[ w_{h,i} > 0 ] \not\in [q, 1-q]$$
$$\Pr_{h \in V}[w_{s,i} = 1] \not\in [q, 1-q]$$

Also, by the contradiction assumption,

$$\Pr_{h \in V}[ w_h \cdot (e_0 + e_i) > b_h ] \not\in [q, 1-q]$$
$$\Pr_{h \in V}[ 1 + w_{h,i} > 0 ] \not\in [q, 1-q]$$
$$\Pr_{h \in V}[w_{h,i} \neq -1] \not\in [q, 1-q]$$
$$\Pr_{h \in V}[w_{h,i} = -1] \not\in [q, 1-q]$$

Since $\Pr_{h \in V}[w_{h,i} = 1] + \Pr_h[w_{h,i} = 0] + \Pr_h[w_{h,i} = 1] = 1$,

$$\Pr_{h \in V}[ w_{h,i} = 0 ] \not\in [q, 1-2q]$$

Thus, each index is either mostly $1$, mostly $0$, or mostly $-1$ for elements in $S$ (since $q<1/3$). Define $m \in \{-1, 0 1\}^d$ such that

$$m_i = \argmax_c \Pr_{h \in V}[w_{h,i} = c]$$

\end{proof}
Note that $m_1 = 1$.

\subsubsection{Ratio between $m^{(-)}$ and $m^{(+)}$}

Note,
$$\mathbb{E}_{h \in V}[w_h^{(+)}] = \sum_{i=1}^d \Pr[w_{h,i} = 1]$$
$$\leq (q) (d - m^{(+)}) + (1) m^{(+)} = qd + (1-q) m^{(+)}$$
$$m^{(+)} \geq \frac{1}{1-q} (\mathbb{E}_{h \in V}[w_h^{(+)}] - qd)$$

Further note,
$$\mathbb{E}_{h \in V}[w_h^{(+)}] = \sum_{i=1}^d \Pr[w_{h,i} = 1]$$
$$\geq (0) (d - m^{(+)}) + (1-q) m^{(+)} = (1-q) m^{(+)}$$
$$m^{(+)} \leq \frac{1}{1-q} \mathbb{E}_{h \in V}[w_h^{(+)}]$$

We have similar equations for $m^{(-)}$ and $\mathbb{E}_{h \in V}[w_h^{(-)}]$

Let $\bar{m}$ be the vector of $m$ without the first component.

Recall that we have

$$\forall h \in V: w_h^{(+)} \leq r w_h^{(-)} - \frac{1}{8}d$$
$$ \mathbb{E}_{h \in V}[w_h^{(+)}] \leq r \mathbb{E}_{h \in V}[w_h^{(-)}] - qrd$$
$$ \frac{1}{1-q} \mathbb{E}_{h \in V}[w_h^{(+)}] \leq r \frac{1}{1-q} (\mathbb{E}_{h \in V}[w_h^{(-)}] - qd)$$
$$m^{(+)} \leq r m^{(-)}$$
$$\bar{m}^{(+)} \leq r \bar{m}^{(-)}$$

Also, recall,

$$\forall h \in V: w_h^{(-)} \leq r (w_h^{(+)} - 1) - \frac{1}{8}d $$
$$ \mathbb{E}_{h \in V}[w_h^{(-)}] \leq r \mathbb{E}_{h \in V}[w_h^{(+)}] - qrd - r$$
$$ \frac{1}{1-q} \mathbb{E}_{h \in V}[w_h^{(-)}] \leq r \frac{1}{1-q} (\mathbb{E}_{h \in V}[w_h^{(+)}] - qd) - \frac{r}{1-q}$$
$$m^{(-)} \leq r m^{(+)} - \frac{r}{1-q}$$
$$m^{(-)} \leq r (m^{(+)} - 1)$$
$$\bar{m}^{(-)} \leq r \bar{m}^{(+)}$$

\subsubsection{Partition}

Let $\bar{w}$ be the vector of $w$ without the first component.

\begin{definition}
Let
\begin{itemize}
\item $\sX^+ = \{x: \Pr_{h \in V}[\bar{w_h} \cdot x \geq 1] > 1 - q\}$
\item $\sX^0 = \{x: \Pr_{h \in V}[\bar{w_h} \cdot x = 0 ] > 1 - 2q\}$
\item $\sX^- = \{x: \Pr_{h \in V}[\bar{w_h} \cdot x \leq -1] > 1 - q\}$
\end{itemize}
\end{definition}

\begin{lemma}
$\sX^+,\sX^0,\sX^-$ is a partition of $\{0,1\}^{d-1}$
\end{lemma}
\begin{proof}
Since $q \leq 1/4$ and the three defining events are mutually exclusive. It is clear that $A^+,A^0,A^-$ are disjoint. Next we show that every point is in at least one of the sets. Suppose a point $x$ is in neither $A^+$ or $A^-$.

Using the contradiction assumption on the test $(0,x)$,

$$\Pr_{h \in V}[w_h \cdot (0,x) > 0] \not\in [q,1-q]$$
$$\Pr_{h \in V}[\bar{w_h} \cdot x > 0] \not\in [q, 1-q]$$
$$\Pr_{h \in V}[\bar{w_h} \cdot x > 0] < q$$

Using the contradiction assumption on the test $(1,x)$,

$$\Pr_{h \in V}[w_h \cdot (1,x) > 0] \not\in [q,1-q]$$
$$\Pr_{h \in V}[1 + \bar{w_h} \cdot x > 0] \not\in [q, 1-q]$$
$$\Pr_{h \in V}[\bar{w_h} \cdot x \geq 0] \not\in [q, 1-q]$$
$$\Pr_{h \in V}[\bar{w_h} \cdot x < 0] < q$$

Combining these,

$$\Pr_{h \in V}[\bar{w_h} \cdot x = 0] = 1 - \Pr_{h \in V}[\bar{w_h} \cdot x > 0] - \Pr_{h \in V}[\bar{w_h} \cdot x < 0]$$
$$> 1 - 2q$$

Thus, $x \not\in \sX^+$ and $x \not\in \sX^-$ imply $x \in \sX^0$ so the three sets are a partition.
\end{proof}

\begin{definition}
Define $\sX^*$ to be every $x \in \{0,1\}^d$ that $(\bar{m} = 1) \cdot x = (\bar{m} = -1) \cdot x$, where $(\bar{m} = 1)$ is the element-wise boolean function. 
\end{definition}
Intuitively, this means that there are as many ones of $x$ in positions where $\bar{m}=1$ as there are places where $\bar{m}=-1$.

\begin{lemma}
$\sX^* \subseteq \sX^0$
\end{lemma}
\begin{proof}
We prove this by induction on the number of $1$'s in $x$ for $x \in \sX^*$. 

The base case is $x = 0^d$ which is trivially in $\sX^0$.

For other $x$, suppose $x_i=1$ at a location where $m_i=0$. Then we know $x - e_i \in \sX^0$ by the induction hypothesis.

$$Pr_{h \in V}[w_h \cdot (x - e_i) = 0] > 1 - 2q$$
$$Pr_{h \in V}[w_{h,i} =0] > 1-2q$$

From these,
$$Pr_{h \in V}[w_h \cdot x = 0] > 1-4q \geq q$$

for $q \leq 1/5$.
So $x \not\in \sX^+ \cup \sX^-$ and thus $x \in \sX^0$.

The only other case is where $x_i=x_j=1$ at locations where $m_i=1$ and $m_j=-1$. Then we know $x - e_i - e_j \in \sX^0$ from the induction hypothesis.

$$Pr_{h \in V}[w_h \cdot (x - e_i - e_j) = 0] > 1 - 2q$$
$$Pr_{h \in V}[w_{h,i} = 1] > 1-q$$
$$Pr_{h \in V}[w_{h,j} = -1] > 1-q$$

From these,
$$Pr_{h \in V}[w_h \cdot x = 0] > 1-4q \geq q$$

and similarly, $x \in \sX^0$.

\end{proof}

\subsubsection{Probability Distribution}
We now define a probability distribution over $x \in \sX^*$. 

Without loss of generality, suppose $\bar{m}^{(+)} \geq \bar{m}^{(-)}$. 
\begin{itemize}
\item Randomly draw an injection $f: \{i: \bar{m}_i=-1\} \rightarrow \{i: \bar{m}_i=1\}$.
\item Initialize $x=0^{d-1}$
\item For indices $\{i: \bar{m}_i\leq 0\}$, draw $x_i \sim \text{bernoulli}(1/2)$. \item For $\{i: \bar{m}_i = -1\}$, set $x_{f(i)} = x_i$
\end{itemize}

Note that the result $x \in \sX^*$ because of the pairing $f$, there will be a $1$ where $\bar{m}_i=1$ for each $1$ where $\bar{m}_i=-1$.

\subsubsection{Set T}

\begin{definition}
For the probability distribution, 
$$Q(h) = \Pr_{x \in \sX^*} [w_h \cdot x = 0] $$
\end{definition}

\begin{lemma}
Let $T = \{h \in V : Q(h) > 1-4q\}$, then $|V| > 5 |T|$.
\end{lemma}
\begin{proof}
For $x \in \sX^*$, since $\sX^* \subseteq \sX^0$,

$$\Pr_{h \in V}[w_h \cdot x  = 0] > 1 - 2q$$
$$\frac{\sum_{h \in V} \mathbf{1}[w_h \cdot x  = 0]}{|V|} > 1 - 2q$$
$$\sum_{x \in \sX^0}  P(x) \frac{\sum_{h \in V} \mathbf{1}[w_h \cdot x  = 0]}{|V|} > 1 - 2q$$
$$\frac{\sum_{h \in V} \sum_{x \in \sX^0}  P(x) \mathbf{1}[w_h \cdot x  = 0]}{|V|} > 1 - 2q$$
$$\frac{\sum_{h \in V} Q(h)}{|V|} > 1 - 2q$$

$$\frac{|T|}{|V|} (1) + \frac{|V| - |T|}{|V|} (1 - 4q) > 1 - 2q$$

$$2 |T| > |V|$$

\end{proof}

\begin{lemma}
$|T| \leq 3$
\end{lemma}
\begin{proof}

Recall that $\bar{m}^{(+)} \leq r \bar{m}^{(-)}$ and $\bar{m}^{(-)} \leq r \bar{m}^{(+)}$ as well

Also $1-4q \geq 1-\min(\frac{1}{4},\frac{1}{2r})$ since $q \leq \min(\frac{1}{16},\frac{1}{8r})$

For any $t \in T$, $Q(t) > 1-4q \geq 1 - \min(\frac{1}{4},\frac{1}{2r})$. Define $\text{Ber}(1/2)$ to be a Bernoulli random variable. 

$$\Pr_{x \in \sX^*} [w_t \cdot x = 0] > 1 - \min(\frac{1}{4},\frac{1}{2r})$$
$$\mathbb{E}_{f} [\Pr[\sum_{i: m_i=0} w_{t,i} \text{Ber}(1/2) + ...$$ $$\sum_{i: m_i=-1} (w_{t,i} + w_{t,f(i)}) \text{Ber}(1/2) = 0]]  > 1 - \min(\frac{1}{4},\frac{1}{2r})$$

Note that

$$\Pr[\sum_{i: \bar{m}_i=0} w_{t,i} \text{Ber}(1/2) + ...$$ $$ \sum_{i: \bar{m}_i=-1} (w_{t,i} + w_{t,f(i)}) \text{Ber}(1/2) = 0] \leq \frac{1}{2}$$

unless $\forall i: \bar{m}_i=0: w_{t,i}=0$ and $\forall i: \bar{m}_i=-1: w_{t,i} + w_{t,f(i)}=0$, call this $condition(t,f)$.

$$\mathbb{E}_{f} [\mathbf{1}[condition(t,f)] + ...$$ $$ \frac{1}{2}(1 - \mathbf{1}[condition(t,f)])]  > 1 - \min(\frac{1}{4},\frac{1}{2r})$$

$$\Pr_{f} [condition(t,f)] > 1 - \min(\frac{1}{2},\frac{1}{r})$$

If $\bar{m}^{(-)} = 0$, then $\bar{m}^{(+)}=0$, and thus $\forall i: \bar{m}_i=0: w_{t,i}=0$ so $t=0^d$ and $|T|=1 \leq 3$.

Note that $\Pr_{f} [condition(t,f)] > 1/2$ implies that $\forall i: m_i=0: w_{t,i}=0$.

\begin{lemma}
If there exists $i,j$ such that $\bar{m}_i = \bar{m}_j = -1$, then $w_{t,i} = w_{t,j}$.
\end{lemma}
\begin{proof}
$\Pr_{f} [condition(t,f)] > \frac{1}{2}$ means that 

$$\Pr_{f} [w_{t,i} = -w_{t,f(i)}] > \frac{1}{2}$$ 
$$\Pr_{f} [w_{t,j} = -w_{t,f(j)}] > \frac{1}{2}$$

so 

$$\frac{ \{l: \bar{m}_l=1 \wedge w_{t,l} = -w_{t,i}\} }{ \{l: \bar{m}_l=-1 \} } > 1/2$$
$$\frac{ \{l: \bar{m}_l=1 \wedge w_{t,l} = -w_{t,j}\} }{ \{l: \bar{m}_l=-1 \} } > 1/2$$

which is only possible if $w_{t,i} = w_{t,j}$.
\end{proof}

Thus, there is some $c \in \{-1,0,1\}$ such that $\forall i: \bar{m}_i=1: w_{t,i}=c$.

$$\Pr_{f} [condition(t,f)] > \frac{1}{r}$$
$$\Pr_{f} [\forall i: \bar{m}_i=-1: w_{t,f(i)} = -c] > 1 - \frac{1}{r}$$
$$1 - \Pr_{f} [\exists i: \bar{m}_i=-1: w_{t,f(i)} \neq -c] > 1 - \frac{1}{r}$$
$$\Pr_{f} [\exists i: \bar{m}_i=-1: w_{t,f(i)} \neq -c] < \frac{1}{r}$$

Suppose $\exists j: \bar{m}_j=1: w_{t,j} \neq -c$,

$$\Pr_{f} [\exists i: \bar{m}_i=-1: f(i)=j] = \frac{1}{r}$$

which is a contradiction. So $\forall j: \bar{m}_j=1: w_{t,j} = -c$.

Thus, $c$ completely determines $t$. Since there are three options for $c$, there are three options for $t$, and $|T| \leq 3$.
\end{proof}

Since $|T| \leq 3$ and $2 |T| \geq |V|$, $|V| \leq 6$. Thus, there is a split of $1/6$ which is a contradiction since $q \leq \frac{1}{8}$. Thus, the lemma is proved. And thus the binary linear classifier problem is split-neighborly.

\end{document}